\documentclass[11pt]{article}

\usepackage[utf8]{inputenc}
\usepackage[T1]{fontenc}

\usepackage[verbose=true,
letterpaper,
textheight=8.4in,
textwidth=6.1in,
margin=1in,
headheight=12pt,
headsep=25pt,
footskip=30pt]{geometry}

\usepackage[toc,page,header]{appendix}
\usepackage{titletoc}

\usepackage{microtype}

\usepackage{amsmath, amssymb, amsfonts, amsxtra, bm}
\usepackage{mathtools}
\usepackage{amsthm}

\usepackage{booktabs}
\usepackage{makecell,multirow} 
\usepackage{nicefrac} 
\usepackage{colortbl}

\usepackage[usenames,dvipsnames]{xcolor}   
\usepackage{graphicx}
\usepackage{epstopdf}

\usepackage[round]{natbib}
\usepackage{hyperref}
\usepackage{url}

\usepackage{paralist}
\usepackage{enumerate}
\usepackage{enumitem}
\usepackage{xspace}
\usepackage{footnote, tablefootnote}

\usepackage{algorithm}
\usepackage{algpseudocode}

\usepackage{comment}
\usepackage{tikz}
\usepackage{tcolorbox}
\usepackage{bbm}

\usepackage{graphicx}
\usepackage{subcaption}

\usepackage{float}
\usepackage{placeins}

\usepackage{microtype}
\usepackage{graphicx}
\usepackage{subcaption}
\usepackage{booktabs} 

\usepackage{hyperref}

\usepackage{algorithm}

\usepackage{placeins}
\usepackage{mdframed} 

\usepackage{amsmath}
\usepackage{amssymb}
\usepackage{mathtools}
\usepackage{amsthm}
\newtheorem*{theorem*}{Theorem}

\usepackage[capitalize,noabbrev]{cleveref}

\theoremstyle{plain}
\newtheorem{theorem}{Theorem}[section]

\newtheorem{lemma}[theorem]{Lemma}

\theoremstyle{definition}

\theoremstyle{remark}
\newtheorem{remark}[theorem]{Remark}

\newenvironment{proofsketch}{\par\noindent\textit{Proof Sketch.}\ }{\hfill$\square$\par}

\newcommand{\bemph}[1]{\textbf{\emph{#1}}}

\titlecontents{section}[3em]{\vspace{-1pt}}%
    {\bfseries\contentslabel{2em}}
    {}
    {\titlerule*[0.5pc]{.}\contentspage}%
    
\titlecontents{subsection}[5em]{\vspace{-1pt}}%
    {\contentslabel{2em}}
    {}
    {\titlerule*[0.5pc]{.}\contentspage}%

\titlecontents{subsubsection}[5em]{\vspace{-1pt}}%
    {\footnotesize\contentslabel{3em}}
    {}
    {\titlerule*[0.5pc]{.}\contentspage}%

\numberwithin{equation}{section}
\usepackage{ss} 

\begin{document}

\title{Toward generalizable learning of all (linear) first-order methods via memory augmented Transformers}

\author{\name{Sanchayan Dutta}\email{dutta@ucdavis.edu}\\
\addr{UC Davis, Davis CA, USA}\\
\name{Suvrit Sra} \email{s.sra@tum.de}\\
\addr{TU Munich, Garching, Germany}
}

\maketitle

\begin{abstract}
We show that memory-augmented Transformers can implement the \bemph{entire class of linear first-order methods} (LFOMs), a class that contains gradient descent (GD) and more advanced methods such as conjugate gradient descent (CGD), momentum methods and all other variants that linearly combine past gradients. Building on prior work that studies how Transformers simulate GD, we provide theoretical and empirical evidence that memory-augmented Transformers can learn more advanced algorithms. We then take a first step toward turning the learned algorithms into actually usable methods by developing a mixture-of-experts (MoE) approach for \bemph{test-time adaptation} to out-of-distribution (OOD) samples. Lastly, we show that LFOMs can themselves be treated as learnable algorithms, whose parameters can be learned from data to attain strong performance.
\end{abstract}

\section{Introduction}
\label{Sect:Introduction}

In-context learning (ICL) allows large language models (LLMs) to generate contextually appropriate outputs based solely on examples and queries provided in a prompt~\citep{brown2020language, liu2021makes, lu2021fantastically, wei2022chain, wu2022self}. This remarkable ability has spurred research into understanding how Transformers can implement algorithms~\citep{achiam2023gpt, touvron2023llama}, with recent studies focusing on they can simulate optimization algorithms~\citep{dai2022can, von2023transformers, garg2022can, akyurek2022learning}. Transformers have been shown to implement gradient-based optimization during their forward pass, such as preconditioned gradient descent for linear regression tasks~\citep{dai2022can, mahankali2023one, ahn2024transformers}. 

Other recent work shows that Transformers can also learn more advanced optimization methods, e.g.,  \citet{fu2023transformers} show that Transformers exhibit convergence rates comparable to Newton's Method for ICL on linear regression. \citet{vladymyrov2024linear} prove that Transformers can learn a variant of gradient descent that approximates second-order methods, such as \(\mathrm{GD}^{++}\), achieving convergence rates similar to Newton's method. These findings lead to the central question of our paper:
\begin{center}
    \emph{Can Transformers efficiently learn more advanced gradient-based optimization methods?}
\end{center}
We address this question by investigating the representational power of Transformers as \emph{algorithm learners}, and with it, contributing to the topic of machine driven algorithm discovery. More specifically, we focus on learning all gradient-based algorithms obtainable by linearly combining past gradients, known as \bemph{Linear First-Order Methods (LFOMs)}~\citep{goh2017why}, wherein \((k+1)\)-st iterate is
\begin{equation}
w^{k+1} = w^0 + \sum\nolimits_{i=0}^k \Gamma_i^k \nabla f(w^i),
\label{LFOM_definition}
\end{equation}
and where \(\{\Gamma_i^k\}_{i=0}^k\) are diagonal matrices. Iteration~\eqref{LFOM_definition} is quite general: it includes as special cases standard methods such as gradient descent (GD), momentum GD, Nesterov's accelerated gradient, conjugate gradient, and in a stochastic version, AdaGrad, ADAM, among others.

Toward showing how Transformers can (efficiently) capture all LFOMs, our key insight is to consider memory-augmented Transformers, known as \emph{Memformers}~\citep{wu2020Memformer, xu2021transformer}, which retain intermediate attention values across layers. This memory enables Memformers to store past gradients, and facilitates efficient mimicking of first-order methods such as conjugate gradient descent and momentum methods, that also use past gradients.

One may ask: \bemph{What does it mean to ``learn'' an optimization algorithm?} For us, it refers to two key aspects: 

1.~\bemph{Expressivity.} The Memformer can perform iterations of LFOMs in its forward pass, under a suitable choice of parameters. Thus, the architecture and parameterization are \emph{sufficiently expressive} to simulate LFOMs.

2.~\bemph{Trainability.} The Memformer's parameters can be trained on random linear regression tasks. Then, using these learned parameters (which are shared across all in-context data samples), the Memformer can execute ``CGD-like'' and ``LFOM-like'' iterations during a forward pass. (\emph{cf.}~Section~\ref{sec:memformers_LFOM})

While far from unconditional learning of GD, the above two aspects of learning follow a large body of work on how Transformers learn GD in context~\citep{garg2022can, akyurek2022learning, von2023transformers, ahn2024transformers, zhang2024trained}. Inspired by these works, and extending the work of \citet{ahn2024transformers}, we study ``learning'' by analyzing the loss landscape for memory-augmented \emph{Linear Transformers} that omit softmax activations~\citep{schlag2021linear, von2023transformers, ahn2024transformers}.

\vspace*{-3pt}
\subsection{Main Contributions}
\vspace*{-3pt}
\begin{enumerate}
\setlength{\itemsep}{0pt}
    \item \bemph{Theoretical justification that Memformers can implement LFOM iterations, including CGD.} We provide a rigorous theoretical framework showing that Memformers, when trained on linear regression tasks, can be configured to perform iterations of any desired LFOM in their forward pass. 

    \item \bemph{Empirical evidence that Memformers learn optimization algorithms.} Through extensive experiments, we demonstrate that a Memformer can learn LFOMs by training on random linear regression tasks. Thereafter, using its learned parameters, it can solve new regression problems using its forward pass. Notably, on several inputs such a Memformer performs competitively against (sometimes even outpeforming) CGD, Nesterov AGM, and momentum GD. 

    \textbf{This finding is potentially significant} because while CGD (and several LFOMs) adapts its step-size and conjugacy parameters \emph{per sample}---which translates into \(2B\) parameters \(\{\alpha, \gamma\}\) per iteration, for a batch of size \(B\)---a Memformer reuses one set of learned weights across the entire batch (see also Remark~\ref{rmk:cgd}).

    \item \bemph{Enhanced performance through multi-headed attention and mixture-of-experts (MoE).} We demonstrate that multi-headed attention improves both in-distribution and out-of-distribution (OOD) performance, with an MoE approach that enables each attention head to specialize in handling different data distributions. Additionally, we explore the performance of LFOMs when their parameters are learned directly from data, thus framing LFOMs as \emph{statistically learnable algorithms}, and showcase their ability to generalize effectively across diverse in-context data samples.
    
    \item \bemph{Connections to test-time adaptation and relevance for LLMs.} Beyond in-context adaptation, we investigate ways to update model parameters \emph{at test time} to address shifting distributions, a topic with growing importance in large language models (LLMs). Recent work suggests that appropriately scaling test-time compute, rather than just model size, can yield stronger performance improvements~\citep{snell2024scaling}, and we link these insights to our MoE-based Memformer approach (Section~\ref{sec:test_time_adaptation}) to demonstrate how gating and memory preconditioners can adapt at inference time to handle various distribution shifts~\citep{sun2020test}.

\end{enumerate}
\textbf{Our objective.}
We would like to emphasize that our main objective is to investigate the potential of Memformers to learn advanced optimization algorithms \emph{in a general sense}, as well as their potential to adapt to OOD optimization problems. We are \emph{not} advocating for Transformers as replacement for established optimization methods. 
Nevertheless, we hope that in the future, machine-discovered optimization algorithms surpass hand-designed methods such as GD or CGD, at least in an average-case / distributional sense. 

\subsection{Related Work}
\textbf{In-Context Learning.} 
The ability of Transformer models to perform in-context learning (ICL) has been extensively studied since its introduction by \citet{brown2020language}. Subsequent works have explored how these models adapt to new tasks without requiring parameter updates \citep{xie2021explanation, von2023uncovering, hahn2023theory, liu2021makes, lu2021fantastically, wei2022chain, wu2022self}.

\noindent\textbf{Gradient-Based Methods in Transformers.} 
\citet{garg2022can} analyze the learning of GD within Transformers, particularly in the context of ICL for linear functions. Empirical studies \citep{garg2022can, akyurek2022learning, von2023transformers} have shown that Transformers can learn GD after being trained on random linear regression tasks. Expanding on these results, \citet{von2023transformers, ahn2024transformers} demonstrate that Transformers can implement preconditioned GD for solving linear regression problems presented in input prompts. Notably, these works, as well as ours, utilize Linear Transformers as discussed in~\citep{schlag2021linear, von2023transformers, ahn2023linear}.

\noindent\textbf{Higher-Order Optimization Methods in Transformers.} 
Transformers have also been shown to learn higher-order optimization techniques, such as Newton's method, expanding their capabilities beyond first-order methods~\citep{fu2023transformers, giannou2024well, vladymyrov2024linear}.

\noindent\textbf{Memory-Augmented Transformers (Memformers).} 
Memformers were introduced in~\citep{wu2020Memformer, xu2021transformer}. While significant progress has been made in understanding how Transformers can learn GD, their potential for learning more sophisticated methods remains largely unexplored. Our work addresses this gap and shows how Memformers can efficiently implement a wide range of advanced first-order and quasi-second-order methods. 

\section{Background and Problem Setup}
\label{Section:LinearTransformerArchitecture}
We follow the setup of training Transformers on random instances of linear regression, following the prior works \citep{garg2022can, akyurek2022learning, von2023transformers, ahn2024transformers}. We largely follow the notation and formal setup of~\citep{ahn2024transformers}, which we recall below.

\subsection{Linear Transformers on Random Linear Regression}

\textbf{Data Distribution.} Let \(\mathbf{x}(i) \in \mathbb{R}^d\) represent covariates drawn independently from a distribution \(\mathcal{D}_{\mathbf{X}}\), and let \(\mathbf{w}^* \in \mathbb{R}^d\) be drawn from \(\mathcal{D}_{\mathbf{W}}\). The matrix of covariates \(\mathbf{X} \in \mathbb{R}^{(n + 1) \times d}\) contains rows \(\mathbf{x}(i)\). The responses are \(\mathbf{y} = [\langle \mathbf{x}(1), \mathbf{w}^* \rangle, \dots, \langle \mathbf{x}(n), \mathbf{w}^* \rangle] \in \mathbb{R}^n\). Define the input matrix \(\mathbf{Z}_0 \in \mathbb{R}^{(d+1) \times (n+1)}\) as:
\begin{equation}
\mathbf{Z}_0 = 
\begin{bmatrix}
\mathbf{x}(1) & \mathbf{x}(2) & \cdots & \mathbf{x}(n) & \mathbf{x}(n+1) \\
\mathbf{y}(1) & \mathbf{y}(2) & \cdots & \mathbf{y}(n) & 0
\end{bmatrix},
\end{equation}
where the zero corresponds to the unknown response for \(\mathbf{x}(n+1)\). The task is to predict \((\mathbf{w}^*)^{\top} \mathbf{x}(n+1)\) using \(\mathbf{Z}_0\). The training data consists of pairs \((\mathbf{Z}_0, (\mathbf{w}^*)^\top \mathbf{x}(n+1))\) for \(\mathbf{x}(i) \sim \mathcal{D}_{\mathbf{X}}\) and \(\mathbf{w}^* \sim \mathcal{D}_{\mathbf{W}}\).

\textbf{Self-Attention Without Softmax.} We focus on the linear self-attention layer, building on \citep{schlag2021linear, von2023transformers}. Let \(\mathbf{Z} \in \mathbb{R}^{(d+1) \times (n+1)}\) be the input matrix of \(n+1\) tokens in \(\mathbb{R}^{d+1}\). Standard self-attention layer is defined as
\begin{equation}
\text{Attn}_{\text{smax}}(\mathbf{Z}) := W_v \mathbf{Z} M \cdot \text{smax}(\mathbf{Z}^{\top} W_k^{\top} W_q \mathbf{Z}),
\end{equation}
where \(W_v, W_k, W_q \in \mathbb{R}^{(d+1) \times (d+1)}\) are weight matrices, and \(\text{smax}(\cdot)\) denotes the column-wise softmax. The masking matrix \(M\) ensures that the label for \(\mathbf{x}(n+1)\) is excluded is given by
\begin{equation}
M = 
\begin{bmatrix}
\mathbf{I}_n & 0 \\
0 & 0
\end{bmatrix} \in \mathbb{R}^{(n+1) \times (n+1)}.
\end{equation}
Omitting softmax, the attention mechanism becomes
\begin{equation}
\label{eq:attention}
\text{Attn}_{P, Q} (\mathbf{Z}) := P \mathbf{Z} M (\mathbf{Z}^{\top} Q \mathbf{Z}),
\end{equation}
where \(P = W_v\) and \(Q = W_k^{\top} W_q\). This simplified form, as shown in \cite{ahn2024transformers}, can implement preconditioned gradient descent, and it is the one we also use.

\textbf{Architecture.} Following the related work, we also simplify the Transformer to consider only attention layers, using \(L\) layers of linear self-attention with a residual connection. Therefore, for each layer \(\ell\), the output is updated as
\begin{equation}
\mathbf{Z}_{\ell+1} = \mathbf{Z}_\ell + \frac{1}{n} \text{Attn}_{P_\ell, Q_\ell} (\mathbf{Z}_\ell), \quad \ell = 0, 1, \dots, L-1.
\label{Z-update}
\end{equation}
Using~\eqref{Z-update}, with the input $\mathbf{Z}_0$, the final Transformer output is
\begin{equation}
\text{TF}_L(\mathbf{Z}_0; \{P_\ell, Q_\ell\}_{\ell=0}^{L-1}) = -[\mathbf{Z}_L]_{(d+1), (n+1)}.
\end{equation}
The set of parameters $\{P_\ell, Q_\ell\}_{\ell=0}^{L-1}$ is then learned by minimizing the following training objective: \begin{equation}
\mathbb{E}_{(\mathbf{Z}_0, \mathbf{w}^*)} \left[\left( \text{TF}_L(\mathbf{Z}_0) + (\mathbf{w}^{*})^{\top} \mathbf{x}(n+1) \right)^2 \right].
\label{in-context_f}
\end{equation}

We will utilize the following lemma from \cite{ahn2024transformers}, which demonstrates that multi-layer Transformers simulate preconditioned gradient descent under suitable parameterization. We have provided the full proof of this lemma in Appendix~A for completeness.
\begin{equation}
P_\ell = 
\begin{bmatrix}
\mathbf B_\ell = 0_{d \times d} & 0 \\
0 & 1
\end{bmatrix},
\quad
Q_\ell = 
-\begin{bmatrix}
\mathbf{A}_\ell & 0 \\
0 & 0
\end{bmatrix}
\label{params_Thm3}
\end{equation}
\begin{lemma}[Lemma 1, \cite{ahn2024transformers}]
    Let an \(L\)-layer linear transformer be parameterized by \(\mathbf{A}_0, \dots, \mathbf{A}_{L-1}\), as in \eqref{params_Thm3}. Let 
    \(y_\ell^{(n+1)} = [\mathbf{Z}_\ell]_{(d+1),(n+1)}\) for \(\ell = 1, \dots, L\); then,
    \begin{equation}
    y_\ell^{(n+1)} = -\langle \mathbf{x}^{(n+1)}, \mathbf{w}_{\ell}^{\mathrm{gd}} \rangle,
    \end{equation}
    where the sequence \(\{\mathbf{w}_{\ell}^{\mathrm{gd}}\}\) is defined as \(\mathbf{w}_{0}^{\mathrm{gd}} = 0\) and for \(\ell = 1, \dots, L-1\):
    \begin{equation}
    \mathbf{w}_{\ell+1}^{\mathrm{gd}} = \mathbf{w}_{\ell}^{\mathrm{gd}} - \mathbf{A}_\ell \nabla R_{\mathbf{w}^*}(\mathbf{w}_{\ell}^{\mathrm{gd}}),
    \end{equation}
    with the empirical least-squares loss (with \(\mathbf{X} := [\mathbf x^{(1)}, \ldots, \mathbf x^{(n)}] \in \mathbb R^{d \times n}\)):
    \begin{equation}
    R_{\mathbf{w}^*}(\mathbf{w}) := \frac{1}{2n} \|\mathbf{X}^\top \mathbf{w} - \mathbf{X}^\top \mathbf{w}^*\|^2. 
    \label{in-context-loss}
    \end{equation}
\label{Lemma 1}
\end{lemma}
\vspace{-5mm}
\textit{\textbf{Note}}. As observed in \S{}C.1 of \cite{ahn2024transformers}, the term \(\mathrm{Attn}_{P_\ell, Q_\ell}(\mathbf{Z}_\ell)\) in \eqref{Z-update} corresponds to the preconditioned gradient \(\mathbf{A}_\ell \nabla R_{\mathbf{w}^*}(\mathbf{w}_\ell^\mathrm{gd})\) of \eqref{in-context-loss} in the update for \(\mathbf{w}_{\ell+1}^\mathrm{gd}\).
\subsection{Linear First-Order Methods (LFOMs)}
\label{sec:lfom}
Linear First-Order Methods (LFOMs) \citep{goh2017why} optimize smooth functions by iteratively updating a parameter vector \(\mathbf{w}\) with current and past gradients. The general update is
\begin{equation}\label{eq:2}
\mathbf{w}^{k+1} = \mathbf{w}^k + \alpha_k \mathbf{d}^k,
\end{equation}
where \(\alpha_k\) is a step size and \(\mathbf{d}^k\) is a direction typically tied to \(\nabla f(\mathbf{w}^k)\). Different LFOMs vary in how \(\mathbf{d}^k\) and \(\alpha_k\) are chosen. For example:
\[\textbf{(GD)}\quad\ \ 
\mathbf{w}^{k+1}
= \mathbf{w}^0 - \alpha \sum\nolimits_{i=0}^{k} \nabla f(\mathbf{w}^i),\]
\[\textbf{(Momentum)}\quad
\mathbf{w}^{k+1}
= \mathbf{w}^0 + \sum\nolimits_{i=0}^{k} \gamma_i^k \nabla f(\mathbf{w}^i).\]
In more advanced LFOMs, the scalars \(\gamma_i^k\) are replaced by diagonal matrices \(\Gamma_i^k\), enabling coordinate-wise scaling.

\textbf{Conjugate Gradient Descent (CGD).}
Specialized for quadratic minimization, CGD generates update directions \(\mathbf{s}_n\) that are conjugate to previous ones, leading to faster convergence than standard gradient descent. It iterates:
\begin{equation}\Delta \mathbf{w}_n = -\nabla f(\mathbf{w}_n), \quad
\gamma_n = \frac{\|\nabla f(\mathbf{w}_n)\|^2}{\|\nabla f(\mathbf{w}_{n-1})\|^2}\label{cgd-iterates},\end{equation}
\[\mathbf{s}_n = \Delta \mathbf{w}_n + \gamma_n \mathbf{s}_{n-1},\]
then chooses a step size \(\alpha_n\) (often via line search) and updates: \(\mathbf{w}_{n+1} = \mathbf{w}_n + \alpha_n \,\mathbf{s}_n.\)

Both momentum GD and CGD are LFOMs. Momentum methods are widely adopted in modern optimization, while CGD converges in at most \(N\) iterations for \(N\)-dimensional quadratics, and is effective for ill-conditioned problems.
\begin{figure}[t]
  \centering
  \includegraphics[width=0.7\linewidth]{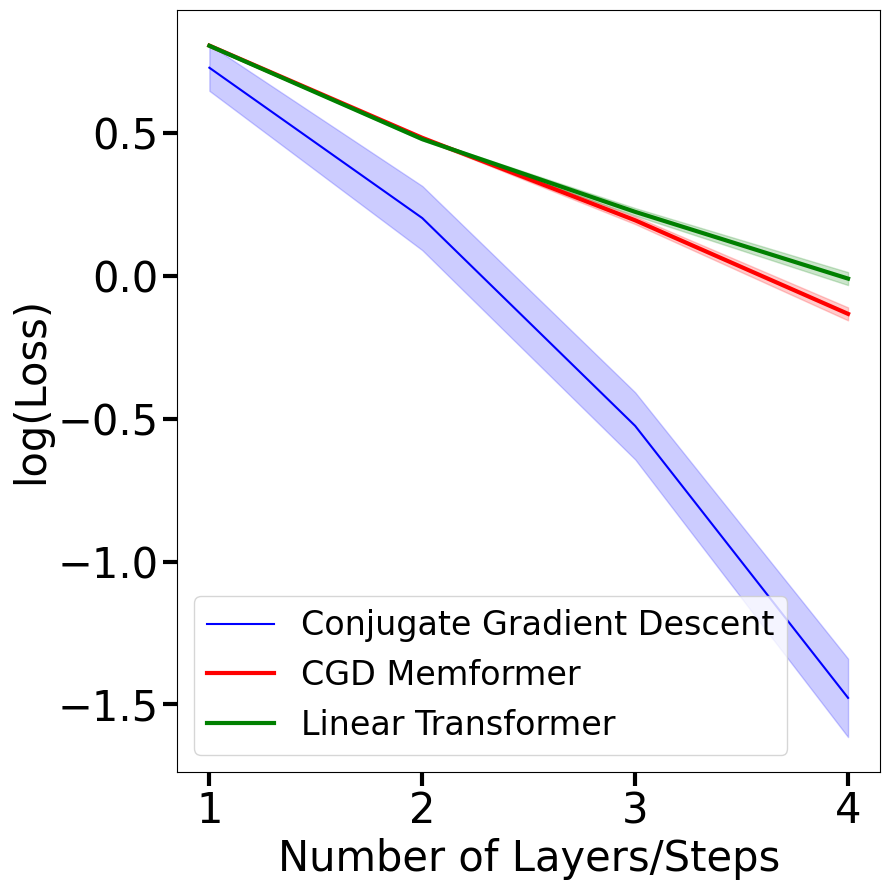}
  \caption{%
    CGD-like Memformer (\ref{eq:dynamic-mem}) \emph{without} preconditioning
    (\(\mathbf{A}_\ell = \mathbf{I}\)) vs. actual CGD running separately on each test sample. Test data is drawn from the same distribution as the training data.
  }
  \label{fig:cg_without_preconditioning}
\end{figure}
\begin{figure}[t]
  \centering
  \includegraphics[width=0.7\linewidth]{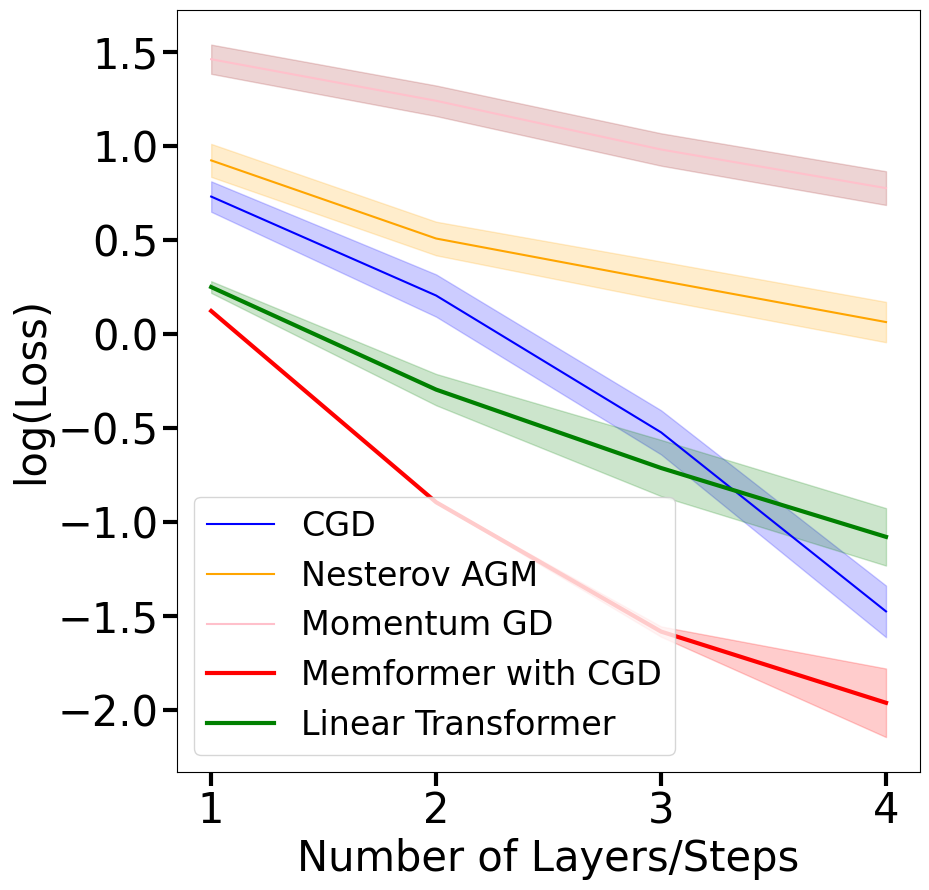}
  \caption{%
    CGD-like Memformer (\ref{eq:dynamic-mem}) \emph{with} preconditioning
    (\(\mathbf{A}_\ell\neq \mathbf{I}\)). This yields a more general LFOM-like scheme,
    often outperforming CGD, Nesterov AGM and momentum GD. Test data is independently drawn from the same distribution as training data.
  }
  \label{fig:cg_with_preconditioning}
\end{figure}
\begin{figure}
  \centering
  \includegraphics[width=0.7\linewidth]{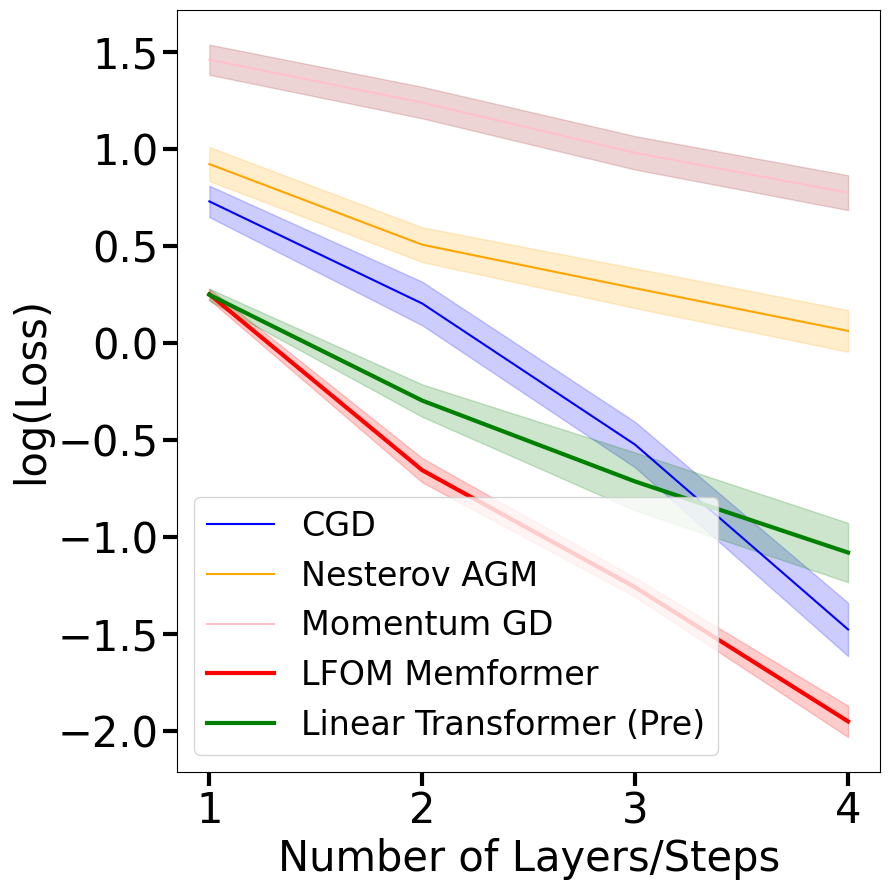}
  \caption{%
    LFOM Memformer (\ref{LFOM_memory}) vs.\ CGD, Nesterov AGM and momentum GD (Pre = non-trivial preconditioners).
  }
  \label{fig:lfom_nonisotropic}
\end{figure}
\section{Memformers Can Implement LFOMs}
\label{sec:memformers_LFOM}
Memformers can ``learn'' LFOMs in the sense described in Section~\ref{Sect:Introduction}. Each layer \(\ell\) has learnable parameters \(\mathbf{A}_\ell, \mathbf{B}_\ell\) \eqref{params_Thm3} and either \(\alpha_\ell, \gamma_\ell\) \eqref{eq:dynamic-mem} or \(\Gamma_\ell\) \eqref{LFOM_memory}. Theorems~\ref{Proposition 1} and~\ref{Proposition 2} show that with suitable parameters, Memformers can \emph{exactly implement} CGD and LFOM iterations in their forward pass.

With only a small number of learned parameters shared across a batch of in-context test data samples, Memformers can perform \textbf{``CGD-like''} (Section~\ref{Sec:dynamic-mem}) or \textbf{``LFOM-like''} (Section~\ref{Sec:LFOM-mem}) updates that rival or even surpass CGD. We refer to algorithms realized by \eqref{eq:dynamic-mem} as ``CGD-like'' and those realized by \eqref{LFOM_memory} as ``LFOM-like''.

\subsection{Single Dynamic Memory for CGD-like Algorithms}
\label{Sec:dynamic-mem}
\begin{theorem}\label{Proposition 1}
A memory-augmented Transformer can implement Conjugate Gradient Descent (CGD) in its forward pass via a single memory register, where:
\begin{align}
    \mathbf{R}_\ell &= \mathrm{Attn}_{P_\ell, Q_\ell}(\mathbf{Z}_\ell) + \gamma_\ell \mathbf{R}_{\ell-1}, \label{eq:dynamic-mem_update}\\
    \mathbf{Z}_{\ell+1} &= \mathbf{Z}_\ell + \alpha_\ell \frac{1}{n} \mathbf{R}_\ell, \label{eq:dynamic-mem}
\end{align}
and \(\gamma_\ell, \alpha_\ell\) control the influence of past updates and step size, respectively.
\end{theorem}
\begin{proofsketch}
Here, \(\mathbf{R}_\ell\) tracks a \emph{single} memory register across layers. CGD updates its search direction by combining the current gradient with the previous direction \eqref{cgd-iterates}. The Transformer mimics this by combining \(\mathrm{Attn}_{P_\ell, Q_\ell}(\mathbf{Z}_\ell)\) (analogous to the gradient) and \(\gamma_\ell \mathbf{R}_{\ell-1}\), and updating the next state \(\mathbf{Z}_{\ell+1}\) via~\eqref{eq:dynamic-mem}. 
With \(\mathbf{A}_\ell = \mathbf{I}\), this matches CGD applied to \eqref{in-context-loss}; see Appendix~A for details. 
\end{proofsketch}

\subsection{Implementing \texorpdfstring{$k$}{k} Steps of LFOM with Separate Memory Registers at Each Layer \(\ell\)}
\label{Sec:LFOM-mem}
Next, we show how Transformers can simulate \(k\) steps of general LFOM updates \eqref{LFOM_definition} by maintaining a separate memory register at each layer \(\ell\).
\begin{theorem}\label{Proposition 2}
A memory-augmented Transformer can implement \(k\) steps of LFOM in its forward pass by maintaining separate memory registers across layers \(\ell\):
\begin{align}
    \mathbf{R}_\ell &= \mathrm{Attn}_{P_\ell, Q_\ell}(\mathbf{Z}_\ell), \\
    \mathbf{Z}_{\ell+1} &= \mathbf{Z}_\ell + \frac{1}{n} \sum_{j=0}^{\ell} \Gamma_j^\ell \odot \mathbf{R}_j, \label{LFOM_memory}
\end{align}
where \(\Gamma_j^\ell\) weights previous layer updates and \(\odot\) is a Hadamard product for scaling.
\footnote{The update~\eqref{LFOM_memory} can be interpreted as a form of \emph{gated memory}. Analogies with LSTMs or GRUs (which also use Hadamard gating) suggest potential ways to refine memory usage in Transformers.}
\end{theorem}
\begin{proofsketch}
Here, \(\mathbf{R}_\ell\) is a \emph{separate} register at each layer. At layer \(\ell\), \(\mathbf{R}_\ell\) captures the current update \(\mathrm{Attn}_{P_\ell, Q_\ell}(\mathbf{Z}_\ell)\). The final output \(\mathbf{Z}_{\ell+1}\) is updated by summing over registers \(\mathbf{R}_j\) scaled through \(\Gamma_j^\ell \in \mathbb R^{(n+1)\times (d+1)}\). This resembles cumulative gradient steps of general LFOMs with diagonal preconditioners~\eqref{LFOM_definition}. A full proof is in Appendix~A.
\end{proofsketch}
The Hadamard product \( \odot \) modulates the influence of \( \mathbf{R}_j \), analogous to gradient preconditioning. This setup subsumes the case of diagonal preconditioners \( \Lambda_i^k \) acting on gradients \( \nabla R_{\mathbf{w}^*}(\mathbf{w}_i^{\mathrm{gd}}) \), which in the general form looks like:
\begin{equation}
\mathbf{w}_{k+1}^{\mathrm{gd}} = \mathbf{w}_0 + \sum_{i=0}^k \Lambda_i^k \nabla R_{\mathbf{w}^*}(\mathbf{w}_i^{\mathrm{gd}}).
\label{lfom_diag_updates}
\end{equation}
The matrices \( \Gamma_j^\ell \in \mathbb{R}^{(d + 1) \times (n + 1)} \) and \( \Lambda_i^k \in \mathbb{R}^{d \times d} \) serve similar roles, but their dimensions differ. \bemph{We expect this memory architecture to be able to perform richer algorithms than LFOMs, though a formal characterization of its full potential remains to be done.}
\noindent

\begin{remark}
\label{rmk:cgd}
Crucially, unlike classical methods such as CGD, which ``adapt'' by computing step sizes (\(\alpha\)) and conjugacy coefficients (\(\gamma\)) separately for each sample—even within the same batch—our Memformer uses a single set of learned parameters shared across \emph{all} samples. For a test batch of size \(B\), CGD thus effectively uses \(2B\) parameters per iteration, whereas the Memformer has just the trained internal parameters \(\{P_\ell, Q_\ell, \Gamma_\ell\}\) \eqref{LFOM_memory}. This amounts to a meta‐optimizer that can exploit “average‐case” behavior over the data distribution, rather than re‐optimizing each instance individually.
\end{remark}

\subsection{Experimental Results}
\label{Section:Experimental_Results}
We evaluate Memformers for learning CGD, general LFOMs, and LFOMs with \(\mathrm{GD}^{++}\) (a quasi-Newton method approximating the Hessian inverse via a truncated Neumann series~\citep{von2023transformers}.

\textbf{Setup.} We use the in-context loss~\eqref{in-context-loss} for linear regression with $d=5$ and $n=20$. Each run samples a random orthogonal matrix $\mathbf{U}\in\mathbb{R}^{d\times d}$ and uses $\mathbf{D} = \mathrm{diag}(1, 1, \tfrac12, \tfrac14, 1)$. With a variance scalar $\sigma^2$ (default $1$), we set $\mathbf{\Sigma} = \sigma^2 \mathbf{U}^\top \mathbf{D} \mathbf{U}$. We draw inputs $\mathbf{x}^{(i)}$ i.i.d.\ from $\mathcal{N}(\mathbf{0}, \mathbf{\Sigma})$, and targets $\mathbf{w}^*$ from $\mathcal{N}(\mathbf{0}, \mathbf{\Sigma}^{-1})$. We optimize the in-context prediction function $f$~\eqref{in-context_f} using a 4-layer linear Transformer trained via ADAM, initializing parameters $\mathbf{A}_0, \mathbf{A}_1, \mathbf{A}_2$, $\mathbf{A}_3$~\eqref{params_Thm3} i.i.d.\ Gaussian, with batch size $1000$ (resampled every 100 steps) and gradient clipping of $0.01$. Plots average over five independent runs with different $\mathbf{U}$.

\textbf{CGD-like Architecture.} 
Figure~\ref{fig:cg_without_preconditioning} shows a Memformer parameterized as in \eqref{eq:dynamic-mem}, where the globally learned \(\alpha_\ell\) and \(\gamma_\ell\) emulate CGD’s line search and deflection. Unlike standard CGD, which refines these parameters individually for each sample, our Memformer uses a single set of learned scalars shared across all data. Despite this approximation, performance remains competitive. Figure~\ref{fig:cg_with_preconditioning} extends the approach by allowing non-scalar preconditioners \(\mathbf{A}_\ell\) \eqref{params_Thm3}, thus making the method a more general ``LFOM-like” strategy that can even outperform standard CGD.

\begin{figure}[t]
  \centering
  \includegraphics[width=0.7\linewidth]{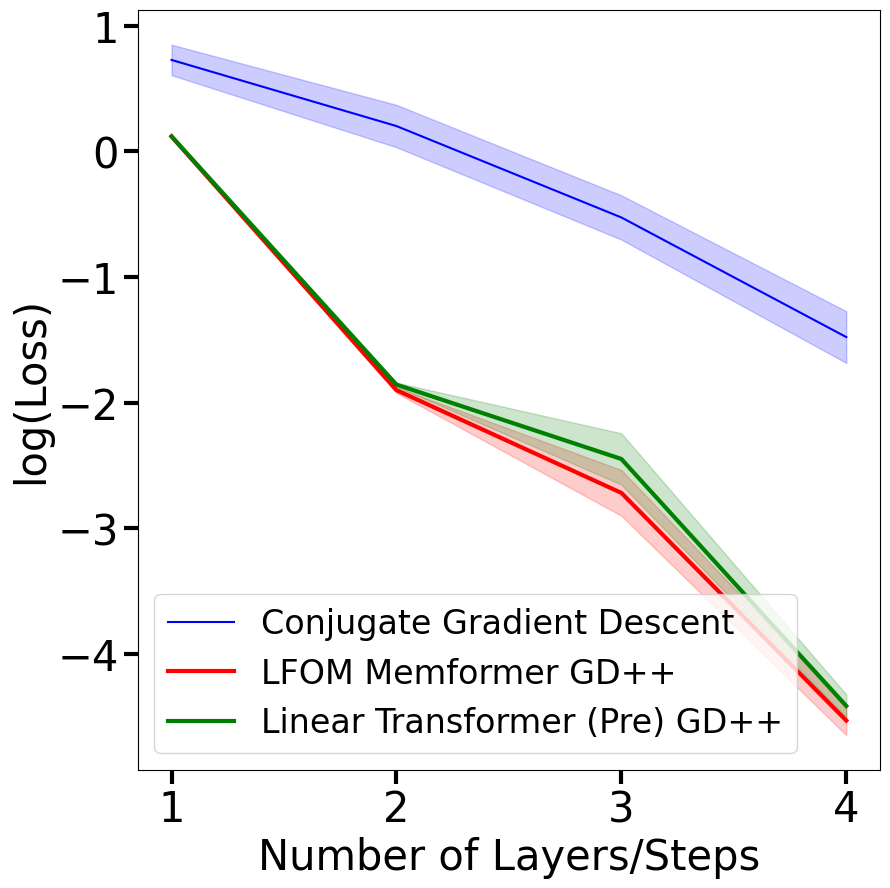}
  \caption{%
    LFOM Memformer with GD++ (\ref{LFOM_memory}) vs. CGD, 
    where the \(\mathbf{B}_\ell\) blocks \eqref{params_Thm3} approximate the Hessian inverse (quasi-Newton).%
  }
  \label{fig:lfom_gdpp_nonisotropic}
\end{figure}

\textbf{LFOM Memformer.}
Figure~\ref{fig:lfom_nonisotropic} shows a Memformer with the LFOM-like architecture \eqref{LFOM_memory}, where each layer’s \(\Gamma_\ell\) acts as a preconditioner. \textbf{In our experiments, we consider the special case of \(\Gamma_j^\ell = \Gamma_j \ \forall \ell\), which is more natural, if we consider that each layer \(j\) of the Memformer has an associated \(\Gamma_j\).} In practice, these parameters further improve performance. Figure~\ref{fig:lfom_gdpp_nonisotropic} extends LFOM with \(\mathrm{GD}^{++}\) by allowing the \(\mathbf{B}_\ell\) blocks \eqref{params_Thm3} to be non-zero. In this case, the \( \mathbf{B}_\ell \) matrices resemble a heavily truncated Neumann series of the inverse \( \mathbf{X} \mathbf{X}^\top \) (Hessian of \eqref{in-context-loss}). Our experiments with more than 4 layers are detailed in Appendix C.
\subsection{Influence of Batch Size on Performance}
\label{Section:Generalized_Performance}
The results in Section~\ref{Section:Experimental_Results} compares the performance of Transformers and Memformers (which learn shared generic parameters during training) against CGD, which computes specific parameters for each observation in a batch of size \( B = 1000 \), independently resampled from the same distribution. While CGD optimizes each observation individually, Transformers and Memformers use shared parameters \( P_\ell, Q_\ell \) (and \(\alpha_\ell, \gamma_\ell\), or \(\Gamma_\ell\)) for each layer \(\ell\), uniformly used for all 1000 observations in the batch. The average log-loss versus layers are plotted for comparison in Figures \ref{fig:cg_with_preconditioning} and \ref{fig:lfom_nonisotropic}. 

\begin{figure}[t]
  \centering
  \includegraphics[width=0.7\linewidth]{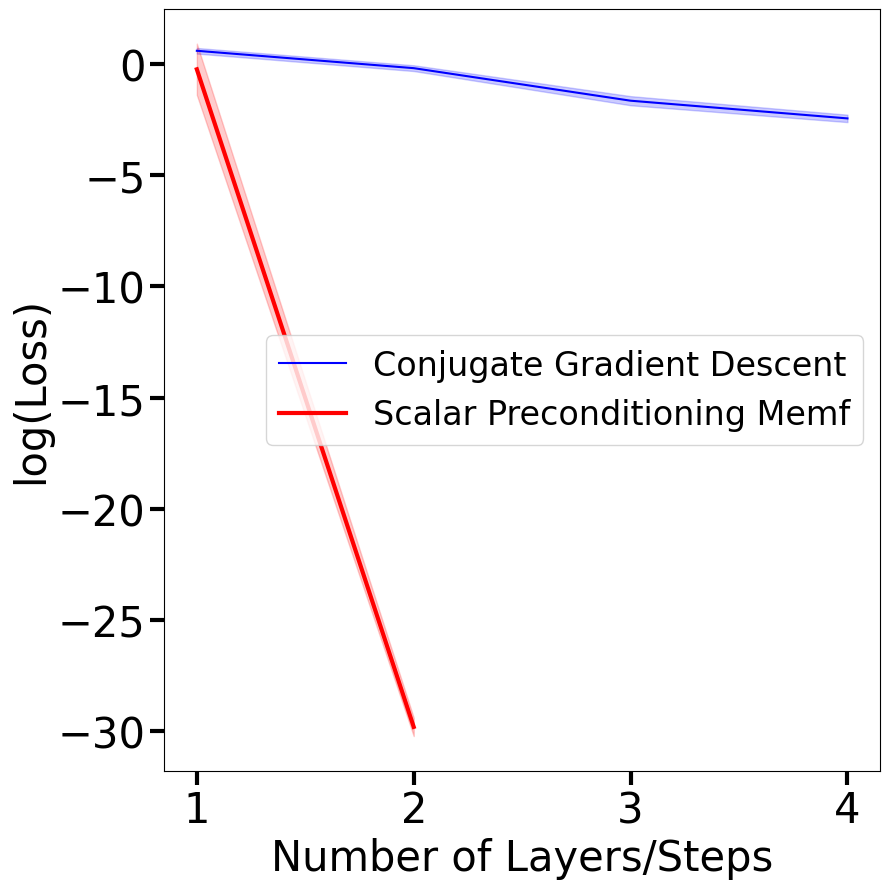}
  \caption{LFOM Memformer \eqref{LFOM_memory} with scalar preconditioners \(\Gamma_\ell\) vs.\ CGD performance on small batch size (\( B = 1 \)). The Memformer demonstrates superior performance.}
  \label{fig:batch_size_1}
\end{figure}

\begin{figure}[t]
  \centering
  \includegraphics[width=0.7\linewidth]{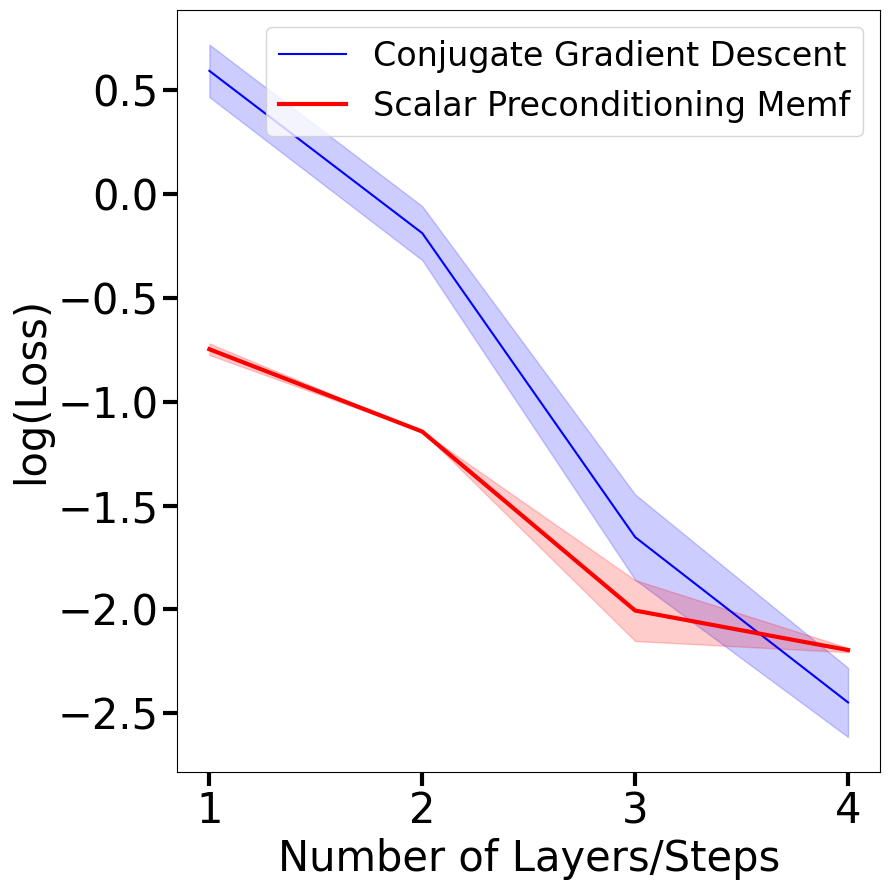}
  \caption{LFOM Memformer \eqref{LFOM_memory} with scalar preconditioners \(\Gamma_\ell\) vs.\ CGD performance on small batch size (\( B = 10 \)). The Memformer outperforms CGD on the training data.}
  \label{fig:batch_size_10}
\end{figure}

The strength of LFOM Memformers \eqref{LFOM_memory} (with matrices \(\Gamma_\ell\) restricted to scalar multiples of the identity) \textbf{becomes even more pronounced when trained on smaller batch sizes, such as \( B = 1 \) and \( B = 10 \)}. In these cases, Memformers learn parameters that significantly outperform CGD, even when CGD runs independently on each observation in the batch. Figures~\ref{fig:batch_size_1} and~\ref{fig:batch_size_10} illustrate this comparison.
\section{Multi-Headed Attention and Out-of-Distribution (OOD) Performance}
\label{Section:Multi-Headed_Attention}
\subsection{OOD Adaptation using Multi-Headed Attention}
Next, we explore how multi-head attention further improves both in-distribution and OOD performance. In our experiments, increasing the number of heads yields notable gains in test-loss metrics. By learning diverse preconditioning matrices, multi-head architectures help Memformers adapt to data with varying covariance structures. Concretely, in the update~\eqref{eq:dynamic-mem_update}, we sum the attention outputs from different heads into a single memory register \(\mathbf{R}_\ell\) at each layer, thus aggregating multiple “views” of the underlying data distribution.
\begin{figure}[t]
  \centering
  \includegraphics[width=0.7\linewidth]{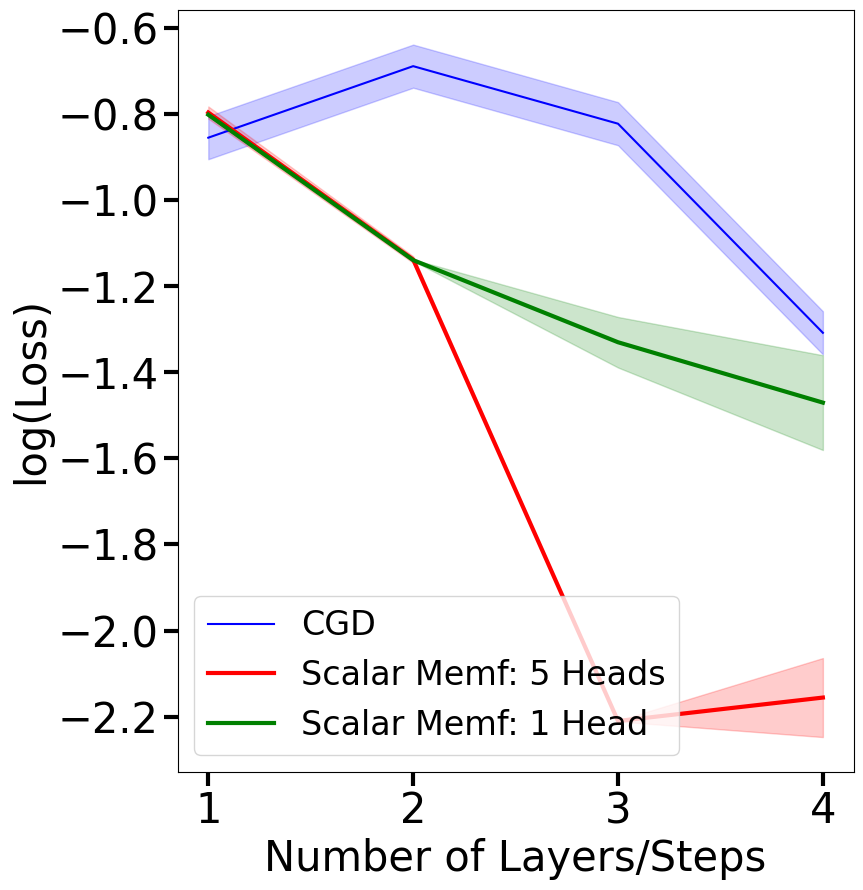}
  \caption{%
  Performance of an LFOM Memformer~\eqref{LFOM_memory} (with scalar
  preconditioners \(\Gamma_\ell\)) comparing one-head and five-head
  attention, relative to CGD. Multi-head attention accelerates convergence
  and improves test performance.}
  \label{fig:5head_attention_experiment}
\end{figure}
These findings align with prior studies underscoring the critical role of multi-head attention in in-context learning. For example, \citet{chen2024transformers} highlight its importance in sparse linear regression, and \citet{cui2024superiority} demonstrate its theoretical and empirical benefits. As illustrated in Figure~\ref{fig:5head_attention_experiment}, the number of heads strongly influences both convergence speed and final test loss.

\subsection{Test-Time Adaptation and Gated Memory}
\label{sec:test_time_adaptation}
Building on the above, we next consider test-time adaptation and show how gating mechanisms allow robust performance under distribution shift.

\begin{figure}[t]
  \centering
  \includegraphics[width=0.7\linewidth]{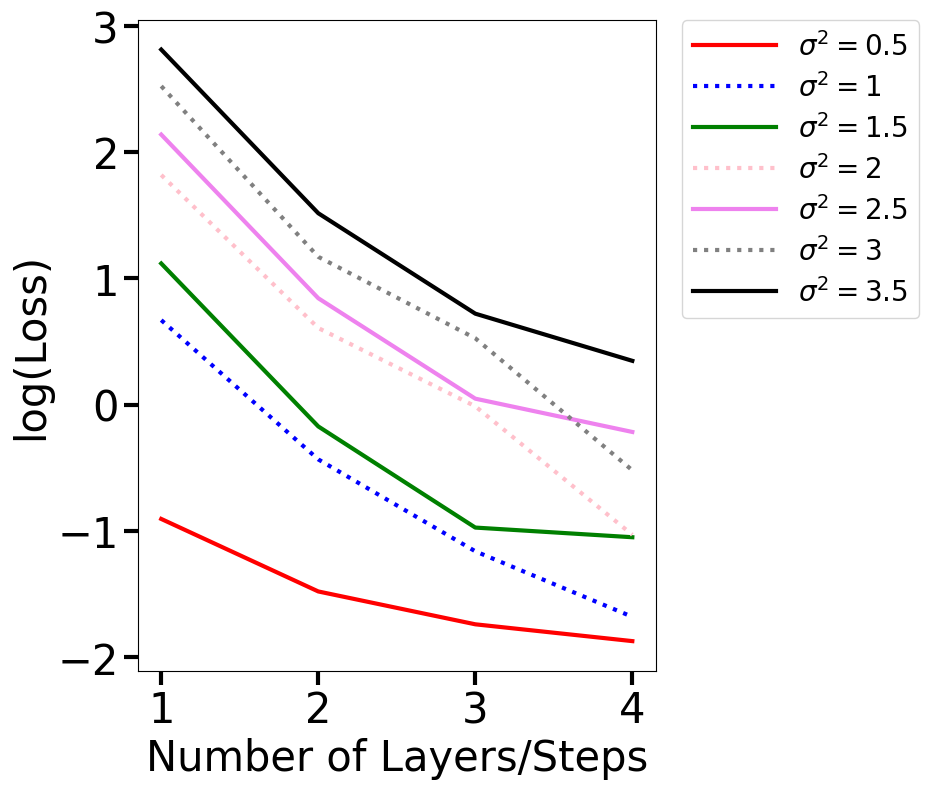}
  \caption{%
  MoE-based 3-head, 4-layer Memformer~\eqref{LFOM_memory} evaluated on
  Gaussian inputs of varying unseen variances. A trainable gating mechanism
  allows effective OOD adaptation.}
  \label{fig:mixture_of_experts_ood}
\end{figure}

\textbf{Mixture of Experts (MoE) for Variance Adaptation.}
To extend multi-head architectures to OOD scenarios, we employ a \emph{mixture of experts (MoE)} approach \cite{jacobs1991adaptive} in a 3-head, 4-layer Memformer. Each head specializes in Gaussian inputs with different variances \(\sigma^2 \in \{1.0, 2.0, 3.0\}\). At inference, a learnable gating mechanism scales the layerwise output \(\sum_{j=0}^{\ell}\Gamma_j \odot \mathbf{R}_j\) \eqref{LFOM_memory} per head coordinate-wise. Concretely, each head \(H_i\) has a scalar gating coefficient \(\alpha_i \in \mathbb{R}\), enabling the model to adapt effectively to unseen variances (e.g., \(\sigma^2 = \{0.5,1.5,3.5\}\)). As shown in Figure~\ref{fig:mixture_of_experts_ood}, this design ensures robust generalization even when the variance differs from the training set.

\textbf{Gaussian Mixture Models (GMMs).}
We further examine OOD adaptation in a 3-head, 4-layer Memformer tested on a Gaussian Mixture Model (GMM) with different means (\(\{2.5,0.0,-1.0\}\)), variances (\(\{0.5,1.0,2.0\}\)), and mixture weights (\(\{0.2,0.3,0.5\}\)). Each head specializes in one mixture component. As before, a gating mechanism similarly scales the layerwise output \(\sum_{j=0}^{\ell}\Gamma_j \odot \mathbf{R}_j\) across heads.

\begin{figure}[t]
  \centering
  \includegraphics[width=0.7\linewidth]{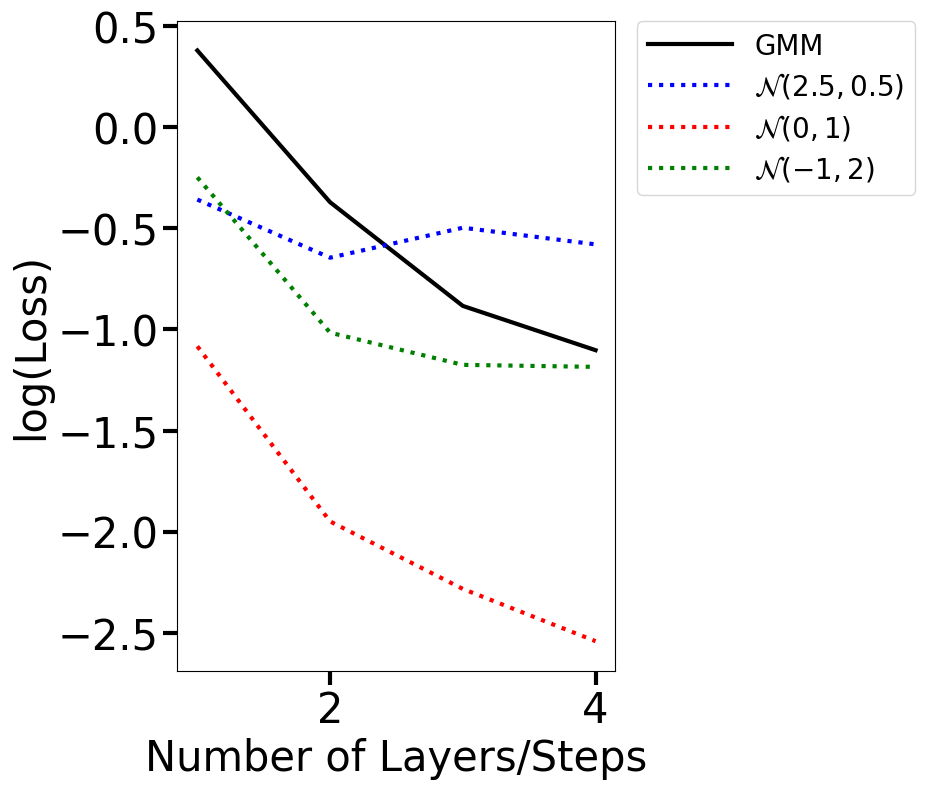}
  \caption{%
  A 3-head, 4-layer Linear Transformer tested on a GMM. Each head specializes
  in one mixture component, and a gating module adjusts the model for OOD data.}
  \label{fig:gaussian_mixture_model_ood_lt}
\end{figure}

\begin{figure}[t]
  \centering
  \includegraphics[width=0.7\linewidth]{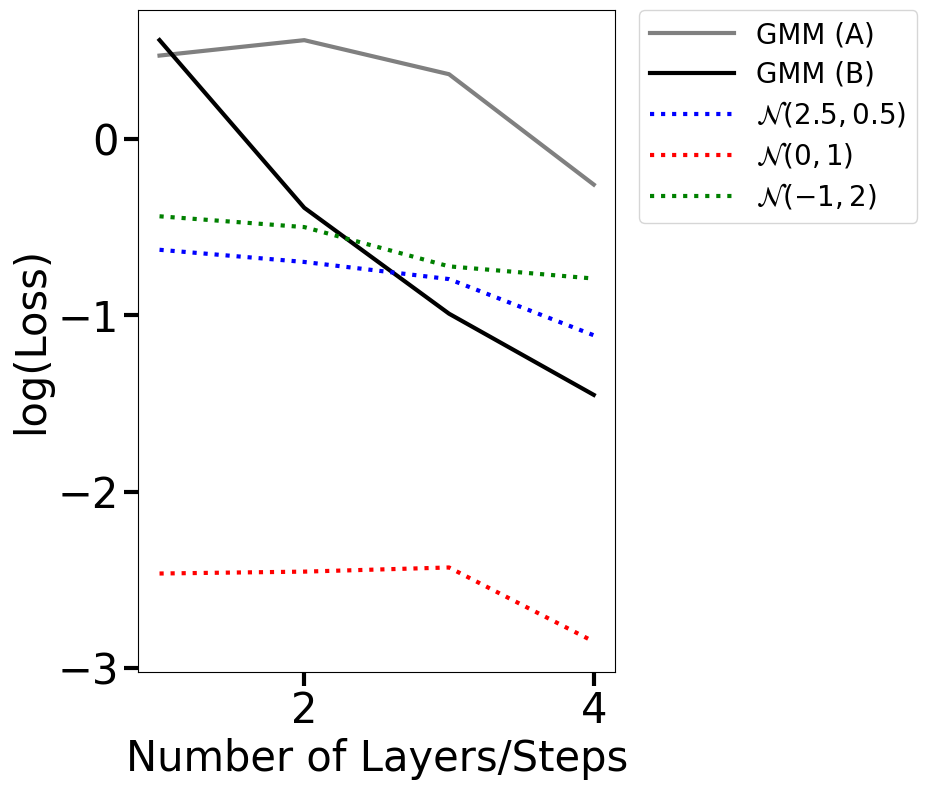}
  \caption{%
  A 3-head, 4-layer Memformer~\eqref{LFOM_memory} tested on a GMM. The gray
  line shows results with a learned gating scheme, while the black line also
  includes test-time updates to the \(\Gamma_j\) matrices.}
  \label{fig:gaussian_mixture_model_ood_mem}
\end{figure}

For this GMM setup, we observe that using a 3-head, 4-layer Linear Transformer (cf. Figure~\ref{fig:gaussian_mixture_model_ood_lt}) gives better performance than a 3-head, 4-layer Memformer (cf. Figure~\ref{fig:gaussian_mixture_model_ood_mem}), as the \(\Gamma_j\) matrices of each head in the Memformer overfit to the individual GMM components. However, this issue can be overcome, with a higher test-time budget, if the \(\Gamma_j\) matrices are also allowed to be trainable at test time (see Figure~\ref{fig:gaussian_mixture_model_ood_mem}).

\begin{theorem}[Multi-Head Memformer with Soft Gating]
\label{thm:multihead_moe_refined}
Consider a multi-head Memformer with $H$ heads, each parameterized by 
$\{P_\ell^h, Q_\ell^h, \Gamma_\ell^h\}$. Suppose $\mathbf{Z}_0$ is drawn from a mixture of $M$ Gaussian components 
$\{\mathcal{N}(\mu_m, \Sigma_m)\}_{m=1}^M$, where each $\mathbf{x}^{(i)} \sim \mathcal{N}(\mu_m, \Sigma_m)$ and $\mathbf{w}^*\sim\mathcal{N}(\mu_m,\Sigma_m^{-1})$. If each component $m$ has at least one head $h_m$ achieving near-optimal first-order performance, then there exist scalar \emph{gating coefficients} $\alpha_{m,h}$ such that for any new prompt from component $m$, the Memformer achieves near-optimal performance via \emph{soft-gating}:
\[
\alpha_{m,1}\,\mathrm{Head}_1 \;+\; \cdots \;+\; \alpha_{m,H}\,\mathrm{Head}_H.
\]
In particular, if $\alpha_{m,h_m} = 1$ and $\alpha_{m,h}=0$ for $h \neq h_m$, the model’s updates align exactly with head $h_m$’s near-optimal method for $\mathcal{N}(\mu_m,\Sigma_m)$.
\footnote{If $\sum_{h=1}^H \alpha_{m,h} = 1$, then $\alpha_{m,1}\,\mathrm{Head}_1 + \cdots + \alpha_{m,H}\,\mathrm{Head}_H$ is a convex combination of the $H$ heads. Such linear updates also appear in momentum/AdamW, where $(1 - \eta\lambda)\theta_{t} + \eta\lambda\,(\text{gradient-based step})$ is itself a convex combination.}
\end{theorem}

A formal proof of Theorem~\ref{thm:multihead_moe_refined} appears in Appendix~A. As indicated above, a Transformer can handle OOD data by fine-tuning specific test-time parameters (e.g., the gating coefficients or the diagonal of \(\Gamma_i\), as in Figure~\ref{fig:gaussian_mixture_model_ood_mem}). This design in effect operates as a meta-version of in-context learning: the main architecture parameters remain fixed, while gating and memory preconditioners can adjust to accommodate distribution shifts, depending on the test-time budget.
\section{LFOMs as Learnable Algorithms}
\label{sec:LFOMs_statistical_learning}
Finally, LFOM updates can themselves be \emph{learned} as a global meta-optimizer, bridging the gap between classical algorithms and in-context algorithmic learning. One can directly learn the parameters of a linear first-order method (LFOM) from data, treating it as a global algorithm that generalizes across random in-context samples. The following theorem provides a standard finite-sample guarantee for such an approach under mild regularity conditions (boundedness, sub-Gaussianity, continuity, and compactness). A complete version and proof appear in Appendix~A.
\begin{theorem}[Statistical Learnability of LFOMs in the In-Context Setting]
\label{thm:lfom_incontext_learnable}
Fix a class of LFOMs parameterized by $\theta\in\Theta\subset\mathbb{R}^p$, e.g., via diagonal preconditioners. Each LFOM $\theta$ maps an in-context prompt $\mathbf{Z}_0$ to a final prediction $\hat{y}_{\theta}(\mathbf{Z}_0)$ for $\mathbf{x}(n+1)$, incurring the loss $\ell(\theta; \mathbf{Z}_0) = \bigl(\hat{y}_\theta(\mathbf{Z}_0) - \langle \mathbf{x}(n+1), \mathbf{w}^*\rangle\bigr)^2$. 

Drawing $N$ i.i.d.\ prompts $\{\mathbf{Z}_0^{(i)}\}_{i=1}^N$ from $\mathcal{D}$, define $\widehat{\theta} = \arg\min_{\theta\in\Theta} \frac{1}{N}\sum_{i=1}^N \ell\bigl(\theta; \mathbf{Z}_0^{(i)}\bigr)$. Then, with high probability over the sample draw, 
\[
    \mathbb{E}_{\mathbf{Z}_0\sim\mathcal{D}}\bigl[\ell(\widehat{\theta};\,\mathbf{Z}_0)\bigr]
    \;\le\;
    \min_{\theta\in\Theta}
    \mathbb{E}_{\mathbf{Z}_0\sim\mathcal{D}}\bigl[\ell(\theta;\,\mathbf{Z}_0)\bigr]
    \;+\;\epsilon(N,\delta),
\]
where $\epsilon(N,\delta) \to 0$ as $N\to\infty$. In other words, the learned LFOM parameters $\widehat{\theta}$ generalize well on new in-context samples from $\mathcal{D}$.
\end{theorem}

\begin{figure}
    \centering
    \includegraphics[width=0.7\linewidth]{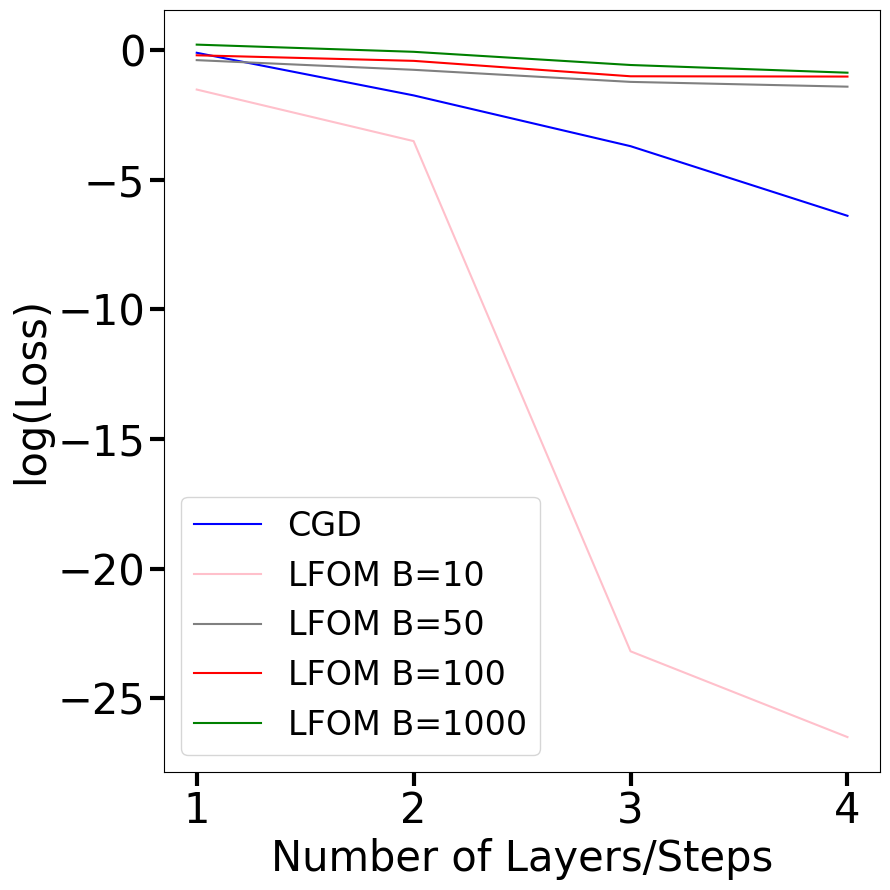}
    \caption{Comparison of a learned LFOM approach vs.\ standard Conjugate Gradient Descent (CGD) on a synthetic quadratic problem \eqref{in-context-loss}. LFOM parameters \eqref{lfom_diag_updates} are trained directly from data, while CGD is run separately on each instance. (\(B\) = Batch Size)}
    \label{fig:ExtractedLFOMvsCGD}
\end{figure}

Note that our Memformer update rule \eqref{LFOM_memory} encompasses an even broader family of optimization strategies than classical LFOMs. Figure~\ref{fig:ExtractedLFOMvsCGD} illustrates how learning LFOM parameters (e.g.\ diagonal preconditioners) directly from data (using ADAM or similar) can match or exceed the performance of Conjugate Gradient Descent (CGD), \emph{while using one shared set of parameters across multiple samples}. Such a strategy can overfit if batches are extremely small, but tends to generalize effectively for moderate to large batch sizes.
\section{Conclusion and Future Work}

We find that memory-augmented Transformers (Memformers) can learn and implement a wide range of first-order optimization algorithms, including gradient descent, conjugate gradient, and momentum methods. This versatility highlights their potential as general-purpose meta-optimizers. Below are several avenues for further exploration:

\textbf{Architectural Extensions, MoE, and TTA.}
Additional gating strategies, mixture-of-experts (MoE), and refined memory modules may boost performance on out-of-distribution (OOD) data. Even a small number of experts can deliver robust gains, though identifying the “right” number of experts remains open. More advanced test-time adaptation (TTA), such as fine-tuning the gating or expert parameters on new data, could further enhance generalization to shifting distributions.

\textbf{Beyond Quadratic Objectives.}
Most theoretical work on algorithm learning in Transformers has focused on linear regression, leaving open the understanding of more complex, potentially nonconvex objectives. By integrating nonlinear attention layers or MLP components, Memformers could tackle tasks like logistic regression or PDE-based inverse problems, where simple quadratic assumptions are insufficient. Analyzing Memformers’ in-context learning on such domains could unify meta-learning research with more realistic, large-scale optimization problems, while clarifying theoretical guarantees beyond the linear-quadratic case.

\textbf{Efficiency vs.\ Generalization.}
While attention-based models can be more computationally demanding than methods like momentum or CGD, they exhibit strong generalization across diverse tasks. Systematic studies of runtime and accuracy trade-offs could clarify the conditions under which learned optimizers outperform classical approaches.

\textbf{Meta-Learning and Transfer.}
Memformers naturally connect to meta-learning by reusing learned optimization strategies. Future investigations might examine more sophisticated TTA, gating, or lightweight fine-tuning approaches to handle distribution shifts. Understanding how best to structure MoE—such as how many experts to deploy and how to select them on the fly—promises both practical efficiency and robust transfer in a variety of domains.

\paragraph{Limitations of our framework.}
While Memformers are versatile, our experiments (Figures \ref{fig:cg_with_preconditioning}, \ref{fig:lfom_nonisotropic}) show they do not drastically outperform advanced methods on quadratic tasks \eqref{in-context-loss}, likely due to the task structure. Future work on broader ICL formulations may shed light on this matter. Notably, Transformers can implement second-order methods like Newton’s \citep{fu2023transformers, giannou2024well}, which typically converge faster and more accurately than LFOMs. However, our focus in this paper is on exploring first-order optimization algorithms that augmented Transformers can learn, rather than proposing a one-size-fits-all optimizer.

\section*{Reproducibility Statement}

We believe the following points provide a clear path for replicating our results:

\begin{itemize}
    \item \textbf{Code Availability}: The code for our experiments, including Memformers and LFOM implementations, is available at \url{https://anonymous.4open.science/r/ICML-2025-LFOM_Memformer}.
    
    \item \textbf{Experiment Setup}: Detailed descriptions of the training setup, model architecture, parameter initialization, and optimization methods are included in Sections \ref{Section:LinearTransformerArchitecture} and \ref{Section:Experimental_Results}.
    
    \item \textbf{Random Seeds}: Random seeds were fixed across all experiments to ensure consistency, and they are provided in the code repository for replication.
    
    \item \textbf{Hardware Requirements}: All experiments were conducted on NVIDIA T4 GPUs in Google Colab.
\end{itemize}

\bibliographystyle{unsrtnat}
\setlength{\bibsep}{3pt}
\bibliography{custom}

\newpage
\appendix
\onecolumn

\section*{Supplementary Material}
\addcontentsline{toc}{section}{Supplementary Material}
\appendix

\section{Proofs}
\subsection{Proof of Lemma~\ref{Lemma 1}: Equivalence to Preconditioned Gradient Descent}
\label{Proof of Lemma 1}

This proof already exists in the literature, for instance, in Subsection C.1 of \cite{ahn2024transformers}. However, we repeat it here, to make this paper as self-contained as possible.

Consider a set of fixed samples \( \mathbf{x}^{(1)}, \dots, \mathbf{x}^{(n)} \), along with a fixed vector \( \mathbf{w}^* \). Let \( P = \{P_i\}_{i=0}^{k} \) and \( Q = \{Q_i\}_{i=0}^{k} \) represent fixed weights, and let \( \mathbf{Z}_i \) evolve as per equation \eqref{Z-update}. Define \( \mathbf{X}_i \) as the first \( d \) rows of \( \mathbf{Z}_k \) (under equation \eqref{params_Thm3}, we have \( \mathbf{X}_i = \mathbf{X}_0 \) for all \( i \)), and let \( \mathbf{Y}_i \) be the \( (d+1) \)-th row of \( \mathbf{Z}_i \). Now, let \( g(\mathbf{x}, \mathbf{y}, k) : \mathbb{R}^d \times \mathbb{R} \times \mathbb{Z} \to \mathbb{R} \) be a function such that for \( \mathbf{x}_{n+1} = \mathbf{x} \) and \( \mathbf{y}_{n+1}^{(0)} = \mathbf{y} \), the function is defined as \( g(\mathbf{x}, \mathbf{y}, k) := \mathbf{y}_{n+1}^{(k)} \). It’s worth noting that \( \mathbf{y}_{n+1}^{(k)} = [\mathbf{Y}_k]_{n+1} \).

We can verify that, under equation \eqref{params_Thm3}, the update rule for \( \mathbf{y}_{n+1}^{(k)} \) is given by:
\begin{equation}
\mathbf{Y}_{k+1} = \mathbf{Y}_k - \frac{1}{n} \mathbf{Y}_k M \mathbf{X}_0^\top A_k \mathbf{X}_0,
\label{Yk_update}
\end{equation}
where \( M \) is a mask matrix of the form:
\[
M = \begin{bmatrix} I & 0 \\ 0 & 0 \end{bmatrix}.
\]

The following points can be verified:

1. \( g(\mathbf{x}, \mathbf{y}, k) = g(\mathbf{x}, 0, k) + \mathbf{y} \). To see this, note that for each \( i \in \{1, \dots, n\} \), we have:
\[
\mathbf{y}^{(i)}_{k+1} = \mathbf{y}^{(i)}_k - \frac{1}{n} \sum_{j=1}^{n} \mathbf{x}^{(i)\top} A_k \mathbf{x}^{(j)} \mathbf{y}^{(j)}_k.
\]
Thus, \( \mathbf{y}^{(i)}_k \) does not depend on \( \mathbf{y}_{n+1}^{(t)} \) for any \( t \). For \( \mathbf{y}_{n+1}^{(k)} \), the update becomes:
\[
\mathbf{y}_{n+1}^{(k+1)} = \mathbf{y}_{n+1}^{(k)} - \frac{1}{n} \sum_{j=1}^{n} \mathbf{x}_{n+1}^\top A_k \mathbf{x}^{(j)} \mathbf{y}^{(j)}_k,
\]
which clearly shows that the dependence on \( \mathbf{y}_{n+1}^{(k)} \) is additive. Through a simple induction, we can establish:
\[
g(\mathbf{x}, \mathbf{y}, k+1) - \mathbf{y} = g(\mathbf{x}, \mathbf{y}, k) - \mathbf{y}.
\]

2. The function \( g(\mathbf{x}, 0, k) \) is linear in \( \mathbf{x} \). To see this, note that for \( j \neq n+1 \), \( \mathbf{y}_j^{(k)} \) does not depend on \( \mathbf{x}_{n+1}^{(t)} \) for any \( t \), \( j \), or \( k \). Therefore, the update for \( \mathbf{y}_{n+1}^{(k+1)} \) depends linearly on \( \mathbf{x}_{n+1} \) and \( \mathbf{y}_{n+1}^{(k)} \). Since \( \mathbf{y}_{n+1}^{(0)} = 0 \) is linear in \( \mathbf{x} \), we conclude by induction that the result holds.

Considering these points, we can confirm that for each \( k \), there exists a vector \( \theta_k \in \mathbb{R}^d \) such that:
\[
g(\mathbf{x}, \mathbf{y}, k) = g(\mathbf{x}, 0, k) + \mathbf{y} = \langle \theta_k, \mathbf{x} \rangle + \mathbf{y},
\]
for all \( \mathbf{x} \) and \( \mathbf{y} \). It follows that \( g(\mathbf{x}, \mathbf{y}, 0) = \mathbf{y} \), so that \( \langle \theta_0, \mathbf{x} \rangle = g(\mathbf{x}, \mathbf{y}, 0) - \mathbf{y} = 0 \), implying \( \theta_0 = 0 \).

We now focus on the third key fact: for each \( i \), we have:
\[
g(\mathbf{x}^{(i)}, \mathbf{y}^{(i)}, k) = \mathbf{y}^{(i)}_k = \langle \theta_k, \mathbf{x}^{(i)} \rangle + \mathbf{y}^{(i)}.
\]
To prove this, let \( \mathbf{x}_{n+1} := \mathbf{x}^{(i)} \) for some \( i \in \{1, \dots, n\} \). Then:
\[
\mathbf{y}^{(i)}_{k+1} = \mathbf{y}^{(i)}_k - \frac{1}{n} \sum_{j=1}^{n} \mathbf{x}^{(i)\top} A_k \mathbf{x}^{(j)} \mathbf{y}^{(j)}_k,
\]
\[
\mathbf{y}_{n+1}^{(k+1)} = \mathbf{y}_{n+1}^{(k)} - \frac{1}{n} \sum_{j=1}^{n} \mathbf{x}_{n+1}^\top A_k \mathbf{x}^{(j)} \mathbf{y}^{(j)}_k,
\]
therefore, \( \mathbf{y}^{(i)}_{k+1} = \mathbf{y}_{n+1}^{(k+1)} \) when \( \mathbf{y}^{(i)}_k = \mathbf{y}_{n+1}^{(k)} \). This completes the induction, given that \( \mathbf{y}^{(i)}_0 = \mathbf{y}_{n+1}^{(0)} \) by definition.

Let \( \mathbf{\bar{X}} \in \mathbb{R}^{d \times n} \) be the matrix whose columns are \( \mathbf{x}^{(1)}, \dots, \mathbf{x}^{(n)} \), excluding \( \mathbf{x}_{n+1} \), and let \( \mathbf{\bar{Y}}_k \in \mathbb{R}^{1 \times n} \) be the vector of \( \mathbf{y}^{(1)}_k, \dots, \mathbf{y}^{(n)}_k \). It follows that:
\[
\mathbf{\bar{Y}}_k = \mathbf{\bar{Y}}_0 + \theta_k^\top \mathbf{\bar{X}}.
\]

Using this, the update formula for \( \mathbf{y}_{n+1}^{(k)} \) becomes:
\begin{equation}
\mathbf{y}_{n+1}^{(k+1)} = \mathbf{y}_{n+1}^{(k)} - \frac{1}{n} \langle A_k \mathbf{\bar{X}}^\top \mathbf{\bar{Y}}_k, \mathbf{x}_{n+1} \rangle,
\label{yn_update}
\end{equation}
leading to the update:
\begin{equation}
\langle \theta_{k+1}, \mathbf{x}_{n+1} \rangle = \langle \theta_k, \mathbf{x}_{n+1} \rangle - \frac{1}{n} \langle A_k \mathbf{\bar{X}} (\mathbf{\bar{X}}^\top \theta_k + \mathbf{\bar{Y}}_0), \mathbf{x}_{n+1} \rangle.
\label{thetak_xn_update}
\end{equation}

Since \( \mathbf{x}_{n+1} \) is arbitrary, we derive the general update formula:
\begin{equation}
\theta_{k+1} = \theta_k - \frac{1}{n} A_k \mathbf{\bar{X}} \mathbf{\bar{X}}^\top (\theta_k + \mathbf{w}^*).
\label{thetak_update}
\end{equation}

Treating \( A_k \) as a preconditioner, and letting \( f(\theta) := \frac{1}{2n} (\theta + \mathbf{w}^*)^\top \mathbf{\bar{X}} \mathbf{\bar{X}}^\top (\theta + \mathbf{w}^*) \), we can express the update as:
\begin{equation}
\theta_{k+1} = \theta_k - \frac{1}{n} A_k \nabla f(\theta).
\label{theta_k_preconditionedGD}
\end{equation}

Finally, let \( \mathbf{w}_k^{\text{gd}} := -\theta_k \). We can verify that \( f(-\mathbf{w}) = R_{\mathbf{w}^*}(\mathbf{w}) \), implying that:
\begin{equation}
\mathbf{w}_{k+1}^{\text{gd}} = \mathbf{w}_k^{\text{gd}} - \frac{1}{n} A_k \nabla R_{\mathbf{w}^*}(\mathbf{w}_k^{\text{gd}}).
\label{wk_update}
\end{equation}

We also confirm that for any \( \mathbf{x}_{n+1} \), the prediction of \( \mathbf{y}_{n+1}^{(k)} \) is:
\[
g(\mathbf{x}_{n+1}, \mathbf{y}_{n+1}, k) = \mathbf{y}_{n+1} - \langle \theta, \mathbf{x}_{n+1} \rangle = \mathbf{y}_{n+1} + \langle \mathbf{w}_k^{\text{gd}}, \mathbf{x}_{n+1} \rangle.
\]

This concludes the proof. We have simply followed the update rule \eqref{Z-update} to its logical conclusion.

\subsection{Full Proof of Theorem~\ref{Proposition 1}}

\begin{theorem*}
A memory-augmented Transformer can implement Conjugate Gradient Descent (CGD) through a dynamic memory mechanism that recursively refines search directions, where the update rules are:
\begin{align}
    \mathbf{R}_\ell &= \mathrm{Attn}_{P_\ell, Q_\ell}(\mathbf{Z}_\ell) + \gamma_\ell \mathbf{R}_{\ell-1}, \label{eq:dynamic-mem_update_proof} \\
    \mathbf{Z}_{\ell+1} &= \mathbf{Z}_\ell + \alpha_\ell \frac{1}{n} \mathbf{R}_\ell, \label{eq:dynamic-mem_proof}
\end{align}
where \( \gamma_\ell \) and \( \alpha_\ell \) control past update influence and step size.
\end{theorem*}

\subsubsection*{Proof}
Our goal is to demonstrate that, under appropriate parameter configurations, the memory-augmented Transformer updates given by equations \eqref{eq:dynamic-mem_update_proof} and \eqref{eq:dynamic-mem_proof} correspond precisely to the Conjugate Gradient Descent (CGD) algorithm when applied to the quadratic loss function:
\begin{equation}
    R_{\mathbf{w}^*}(\mathbf{w}) = \frac{1}{2n} (\mathbf{w} - \mathbf{w}^*)^\top \mathbf{X} \mathbf{X}^\top (\mathbf{w} - \mathbf{w}^*). \label{eq:quadratic_loss}
\end{equation}

We will establish a mapping between the Transformer's operations and the steps of the CGD algorithm, demonstrating that the Transformer can implement CGD under certain parameter settings.

\subsubsection*{CGD Algorithm for Quadratic Functions}
For minimizing a quadratic function, the CGD algorithm proceeds as follows:

\begin{mdframed}[roundcorner=5pt, backgroundcolor=gray!10] 
\textbf{Algorithm: Conjugate Gradient Descent (CGD)}
\begin{itemize}
    \item \textbf{Initialize:} \( \mathbf{w}_0 \), \( \mathbf{r}_0 = -\nabla f(\mathbf{w}_0) \), \( \mathbf{s}_0 = \mathbf{r}_0 \)
    \item \textbf{For} \( n = 1, 2, \dots \):
    \begin{itemize}
        \item Compute the residual: 
        \[
        \mathbf{r}_n = -\nabla f(\mathbf{w}_n)
        \]
        \item Compute the conjugacy coefficient:
        \[
        \gamma_n = \frac{\mathbf{r}_n^\top \mathbf{r}_n}{\mathbf{r}_{n-1}^\top \mathbf{r}_{n-1}}
        \]
        \item Update the search direction:
        \[
        \mathbf{s}_n = \mathbf{r}_n + \gamma_n \mathbf{s}_{n-1}
        \]
        \item Compute the step size:
        \[
        \alpha_n = \frac{\mathbf{r}_n^\top \mathbf{r}_n}{\mathbf{s}_n^\top \mathbf{H} \mathbf{s}_n}
        \]
        \item Update the parameters:
        \[
        \mathbf{w}_{n+1} = \mathbf{w}_n + \alpha_n \mathbf{s}_n
        \]
    \end{itemize}
    \item \textbf{EndFor}
\end{itemize}
\end{mdframed}

\FloatBarrier

\subsubsection*{Mapping CGD Updates to Transformer Updates}

We first recall that in the proof of Lemma 1 (\ref{Proof of Lemma 1}), the \(\mathbf{w}_{k+1}^{\text{gd}}\) update rule
\begin{equation}
\mathbf{w}_{k+1}^{\text{gd}} = \mathbf{w}_k^{\text{gd}} - \frac{1}{n} A_k \nabla R_{\mathbf{w}^*}(\mathbf{w}_k^{\text{gd}}), \label{wk_update}
\end{equation}
is a direct downstream consequence of the \(\mathbf{Z}_{\ell+1}\) update rule \eqref{Z-update}
\[
\mathbf{Z}_{\ell+1} = \mathbf{Z}_\ell + \frac{1}{n} \mathrm{Attn}_{P_\ell, Q_\ell} (\mathbf{Z}_\ell), \quad \ell = 0, 1, \dots, L-1,
\]
under the parameterization given in equation \eqref{params_Thm3}. Thus, the \(\mathrm{Attn}_{P_\ell, Q_\ell}\) term in the \(\mathbf Z_\ell\) update equation is, in a precise sense, paralleled by the \(-\frac{1}{n} A_k \nabla R_{\mathbf{w}^*}(\mathbf{w}_k^{\text{gd}})\) term in the \(\mathbf{w}_{k+1}^{\text{gd}}\) update equation \eqref{wk_update}.

\subsubsection*{Step 1: Initialization}
\begin{itemize}
    \item \textbf{CGD:}
    \[
    \mathbf{w}_0 \text{ given}, \quad \mathbf{r}_0 = -\nabla f(\mathbf{w}_0), \quad \mathbf{s}_0 = \mathbf{r}_0.
    \]
    \item \textbf{Transformer:}
    \begin{itemize}
        \item The initial state \( \mathbf{Z}_0 \) in \eqref{Z-update} parallels \( \mathbf{w}_0 \) in \eqref{wk_update}.
        \item The memory register \( \mathbf{R} \) is initialized to \(\mathrm{Attn}_{P_0, Q_0}(\mathbf{Z}_0)\), i.e., \( \mathbf{R}_{0} =  \mathrm{Attn}_{P_0, Q_0}(\mathbf{Z}_0)\), corresponding to \( \mathbf{s}_{0} = \mathbf{r}_0 \).
        \item We set \( \gamma_0 = 0 \), consistent with CGD initialization.
    \end{itemize}
\end{itemize}

\subsubsection*{Step 2: Update Memory Register (Search Direction)}
\begin{itemize}
    \item \textbf{Transformer Memory Update:}
    \[
    \mathbf{R}_\ell = \mathrm{Attn}_{P_\ell, Q_\ell}(\mathbf{Z}_\ell) + \gamma_\ell \mathbf{R}_{\ell-1}.
    \]
    \item \textbf{Correspondence with CGD:}
    \[
    \mathbf{s}_n = \mathbf{r}_n + \gamma_n \mathbf{s}_{n-1}.
    \]
    Identifying \( \mathbf{R}_\ell \leftrightarrow \mathbf{s}_n \), \( \gamma_\ell = \gamma_n \), and \( \mathbf{R}_{\ell-1} \leftrightarrow \mathbf{s}_{n-1} \), the Transformer's memory update matches CGD.
\end{itemize}

\subsubsection*{Step 3: Update Parameters}
\begin{itemize}
    \item \textbf{Transformer Parameter Update:}
    \[
    \mathbf{Z}_{\ell+1} = \mathbf{Z}_\ell + \alpha_\ell \frac{1}{n} \mathbf{R}_\ell.
    \]
    \item \textbf{Correspondence with CGD:}
    \[
    \mathbf{w}_{n+1} = \mathbf{w}_n + \alpha_n \mathbf{s}_n.
    \]
    The scaling factor \( \frac{1}{n} \) accounts for the gradient's scaling, consistent with the CGD update when considering the Hessian \( \mathbf{H} = \frac{1}{n} \mathbf{X} \mathbf{X}^\top \).
\end{itemize}

\subsubsection*{Step 4: Conjugacy Coefficient \(\gamma_\ell\) and Step Size \(\alpha_\ell\)}
\begin{itemize}
    \item \textbf{CGD Computations}: Scalar values computed based on residuals and the Hessian.
    \item \textbf{Transformer Implementation}: 
    \begin{itemize}
        \item \( \gamma_\ell \) and \( \alpha_\ell \) are treated as parameters, ensuring structural correspondence.
        \item The Transformer's architecture allows these as fixed or learnable parameters.
    \end{itemize}
\end{itemize}

Therefore, under suitable parameter configurations, the memory-augmented Transformer can implement CGD, demonstrating the feasibility of using the Transformer's architecture to perform CGD-like updates.

\subsection{Full Proof of Theorem~\ref{Proposition 2}}

\begin{theorem*}
A memory-augmented Transformer can implement \( k \) steps of Linear First-Order Methods (LFOMs) by maintaining memory registers across layers, where the update rules are:
    \begin{equation}
        \mathbf{R}_\ell = \mathrm{Attn}_{P_\ell, Q_\ell}(\mathbf{Z}_\ell), \label{Memformer_R_update}
    \end{equation}
    \begin{equation}
        \mathbf{Z}_{\ell+1} = \mathbf{Z}_\ell + \frac{1}{n} \sum_{j=0}^{\ell} \Gamma_j^\ell \odot \mathbf{R}_j, \label{Memformer_Z_update}
    \end{equation}
    where \( \Gamma_j^\ell \) governs the contribution of previous layers, and \( \odot \) is the Hadamard (element-wise) product for scaling.
\end{theorem*}

Our goal is to show that the memory-augmented Transformer with updates given by equations \eqref{Memformer_R_update} and \eqref{Memformer_Z_update} can implement \( k \) steps of an LFOM, whose general formulation is:
\[
    \mathbf{w}^{k+1} = \mathbf{w}^0 + \sum_{i=0}^k \Lambda_i^k \nabla f(\mathbf{w}^i), \label{eq:LFOM_update}
\]
where \(\Lambda_i^k\) are diagonal matrices that scale the gradients \(\nabla f(\mathbf{w}^i)\).

We will proceed by establishing a correspondence between the variables and updates in the memory-augmented Transformer and those in the LFOM, and by showing that, under appropriate parameter settings, the Transformer updates replicate the LFOM updates.

The first order of business is to realize that, in the proof of Lemma 1 (\ref{Proof of Lemma 1}), the \(\mathbf{w}_{k+1}^{\text{gd}}\) update rule \eqref{wk_update} is a direct downstream consequence of the \(\mathbf{Z}_{\ell+1}\) update rule \eqref{Z-update}, under the parameterization given in equation \eqref{params_Thm3}.

Set \(\mathbf{R}_\ell = \mathrm{Attn}_{P_\ell, Q_\ell}(\mathbf{Z}_\ell)\) per \eqref{Memformer_R_update}. Then the consequence of the \(\mathbf{Z}_{\ell+1} = \mathbf{Z}_\ell + \frac{1}{n} \sum_{j=0}^{\ell} \Gamma_j^\ell \odot \mathbf{R}_j\) update rule is that each \(\mathrm{Attn}_{P_j, Q_j}(\mathbf{Z}_j)\) is coordinate-wise scaled by \(\Gamma_j^\ell \in \mathbb{R}^{(d + 1) \times (n + 1)}\). But if \(\mathrm{Attn}_{P_j, Q_j}(\mathbf{Z}_j)\) is coordinate-wise scaled by \(\Gamma_j^\ell\), then the \(\mathbf{Y}_{k+1}\) update rule in \eqref{Yk_update} now instead looks like \(\mathbf{Y}_{k+1} = \mathbf{Y}_k - \frac{1}{n} \sum_{j=0}^{k} \Gamma_j^k \big|_{d+1} \odot (\mathbf{Y}_k M \mathbf{X}_0^\top A_k \mathbf{X}_0)\), where \(\Gamma_j^k \big|_{d+1}\) denotes the \((d + 1)\)-th row of \(\Gamma_j^k\). This is because, by definition, \(\mathbf{Y}_i\) is the \((d+1)\)-th row of \(\mathbf{Z}_i\) (\ref{Proof of Lemma 1}).

From the basic \(\mathbf{Y}_k\) update rule in \eqref{Yk_update}, the update formula for \(\mathbf{y}_{n+1}^{(k+1)}\) in \eqref{yn_update} follows as a consequence. Except that now, this update formula will include a coordinate-wise scaling as well, which we will denote by \(\Lambda_j^k \in \mathbb{R}^{d}\):
\[
\mathbf{y}_{n+1}^{(k+1)} = \mathbf{y}_{n+1}^{(k)} - \frac{1}{n} \sum_{j=0}^{k}\langle (A_j \mathbf{\bar{X}}^\top \mathbf{\bar{Y}}_j) \odot \Lambda_j^k, \mathbf{x}_{n+1} \rangle,
\label{scaled_yn_update}
\]
which in turn leads to \(\theta_{k+1} = \theta_k - \frac{1}{n} \sum_{j=0}^{k} (A_j \mathbf{\bar{X}} \mathbf{\bar{X}}^\top (\theta_j + \mathbf{w}^*)) \odot \Lambda_j^k\) in place of \eqref{thetak_update} and \(\mathbf{w}_{k+1}^{\text{gd}} = \mathbf{w}_k^{\text{gd}} - \frac{1}{n} \sum_{j=0}^{k} A_j \nabla R_{\mathbf{w}^*}(\mathbf{w}_j^{\text{gd}}) \odot \Lambda_j^k\) in place of \eqref{theta_k_preconditionedGD}. The negative signs can, of course, be incorporated within the \(\Lambda_j^k\)s.

If we simply rewrite \(\Lambda_j^k \in \mathbb R^{d}\) as a diagonal matrix in \(\mathbb R^{d \times d}\), this setup then subsumes the case of diagonal preconditioners \(\Lambda_j^k \in \mathbb R^{d \times d}\) acting on the gradients \( \nabla R_{\mathbf{w}^*}(\mathbf{w}_j^{\mathrm{gd}}) \), which in the general form looks like:
\[
    \mathbf{w}_{k+1}^{\mathrm{gd}} = \mathbf{w}_0 + \sum_{i=0}^k \Lambda_i^k \nabla R_{\mathbf{w}^*}(\mathbf{w}_i^{\mathrm{gd}}).
\]
where \(\Lambda_i^k\) are diagonal matrices.

\textbf{\textit{Note.}} The memory-augmented Transformer performs exactly these updates in the special case when the preconditioners \(A_j\) are scalar multiples of the identity. If the preconditioners \(A_j\) are non-trivial, then this architecture performs \textbf{``LFOM-like"} algorithms that lie in a class richer than LFOMs (\ref{Sec:LFOM-mem}).

\subsection{Full Proof of Theorem~\ref{thm:multihead_moe_refined}}

\begin{theorem*}
Consider a multi-head Memformer with $H$ heads, where each head $h$ is parameterized by
\[\bigl\{P_\ell^h,\; Q_\ell^h,\; \Gamma_\ell^h\bigr\}_{\ell=1}^L.\]
Suppose that in-context prompts $\mathbf{Z}_0$ are drawn from a mixture of $M$ Gaussian components,
\[\bigl\{\mathcal{N}(\mu_m, \Sigma_m)\bigr\}_{m=1}^M.\]
Assume that, after training, for each mixture component $\mathcal{N}(\mu_m,\Sigma_m)$, there exists a head $h_m$ whose parameters yield near-optimal first-order performance on data from that component. Then, for each component $m$, there exist scalar \emph{gating coefficients}
\[\{\alpha_{m,h}\}_{h=1}^H\]
such that, whenever a new prompt $\mathbf{Z}_0$ is drawn from $\mathcal{N}(\mu_m,\Sigma_m)$, where $\mathbf{x}^{(i)} \sim \mathcal{N}(\mu_m, \Sigma_m)$ i.i.d., and targets $\mathbf{w}^*$ from $\mathcal{N}(\mu_m, \Sigma_m^{-1})$, the Memformer can form the linear combination
\[\sum_{h=1}^H \alpha_{m,h}\,\mathrm{Head}_h(\mathbf{Z}_0)\]
to achieve near-optimal performance for that component. In particular, if one sets
\[\alpha_{m,h_m} = 1
\quad \text{and} \quad
\alpha_{m,h}=0 \ \ \text{for all} \ h \neq h_m,\]
then the multi-head Memformer yields the same near-optimal updates as head $h_m$ alone.
\end{theorem*}

\begin{proof}
For each head $h \in \{1,\dots,H\}$, let $\mathrm{Head}_h(\mathbf{Z}_0)$ denote the final update (or prediction) produced \emph{solely} by that head's parameters
\[
\bigl\{P_\ell^h,\; Q_\ell^h,\; \Gamma_\ell^h\bigr\}_{\ell=1}^L
\]
when presented with the in-context prompt $\mathbf{Z}_0$. By assumption, for each mixture component~$m$, there is an associated head $h_m$ that is near-optimal for data drawn from $\mathcal{N}(\mu_m,\Sigma_m)$. 

We now define gating coefficients
\[\boldsymbol{\alpha}_m
\,=\,(\alpha_{m,1},\,\alpha_{m,2},\,\dots,\,\alpha_{m,H}),\]
to form a linear combination of the $H$ heads' outputs. Specifically, in a forward pass on a new prompt $\mathbf{Z}_0$ sampled from $\mathcal{N}(\mu_m,\Sigma_m)$, the multi-head Memformer can produce
\[\sum_{h=1}^H \alpha_{m,h}\,\mathrm{Head}_h(\mathbf{Z}_0).\]
Choosing $\alpha_{m,h_m} = 1$ and $\alpha_{m,h} = 0$ for $h \neq h_m$ ensures that the combined update coincides exactly with $\mathrm{Head}_{h_m}(\mathbf{Z}_0)$. Since $h_m$ is by hypothesis near-optimal on that mixture component, the overall performance on $\mathcal{N}(\mu_m,\Sigma_m)$ is likewise near-optimal. Repeating this argument for each $m$ shows that every mixture component can be matched with the head that specializes in it.
\end{proof}

\subsection{Full Proof of Theorem~\ref{thm:lfom_incontext_learnable}}

\begin{theorem*}
Fix a class of LFOMs parameterized by $\theta \in \Theta \subset \mathbb{R}^p$, for instance via diagonal preconditioners $\{\Lambda_i^k\}$. Let each LFOM $\theta$ map the in-context prompt $\mathbf{Z}_0$ to a final prediction $\hat{y}_\theta(\mathbf{Z}_0)$ for $\mathbf{x}(n+1)$. Define the in-context loss
\[\ell(\theta; \mathbf{Z}_0) \;=\; \Bigl(\hat{y}_\theta(\mathbf{Z}_0) - \langle \mathbf{x}(n+1), \mathbf{w}^*\rangle \Bigr)^2.\]
Suppose:
\begin{enumerate}
    \item[(i)] The domain of $\mathbf{Z}_0$, as drawn from distribution $\mathcal{D}$, is such that the squared loss $\ell(\theta;\mathbf{Z}_0)$ is almost surely bounded by a constant $L>0$ or is sub-Gaussian with scale $\sigma^2$.
    \item[(ii)] The parameter space $\Theta$ is a compact subset of $\mathbb{R}^p$.
    \item[(iii)] The mapping $\theta \mapsto \ell(\theta;\mathbf{Z}_0)$ is continuous (or Lipschitz) for all $\mathbf{Z}_0$ in the support of $\mathcal{D}$.
\end{enumerate}

Draw $N$ i.i.d.\ prompts $\{\mathbf{Z}_0^{(i)}\}_{i=1}^N \sim \mathcal{D}$, and let
\[\widehat{\theta} \;=\; \arg\min_{\theta\in\Theta} \;\;\frac{1}{N}\sum_{i=1}^N \ell\!\bigl(\theta;\,\mathbf{Z}_0^{(i)}\bigr).\]
Then, with probability at least $1-\delta$ over the sample draw,
\[\mathbb{E}_{\mathbf{Z}_0 \sim \mathcal{D}} \bigl[\ell(\widehat{\theta};\,\mathbf{Z}_0)\bigr] \;\;\le\;\; \min_{\theta \in \Theta} \mathbb{E}_{\mathbf{Z}_0 \sim \mathcal{D}} \bigl[\ell(\theta;\,\mathbf{Z}_0)\bigr] \;+\;\epsilon(N,\delta),\]
where $\epsilon(N,\delta)\to 0$ as $N\to \infty$. Thus the learned LFOM parameters $\widehat{\theta}$ generalize well on new prompts from $\mathcal{D}$.
\end{theorem*}

\begin{proof}
Define the population risk $R(\theta) := \mathbb{E}_{\mathbf{Z}_0 \sim \mathcal{D}} [\ell(\theta;\mathbf{Z}_0)]$ and the empirical risk $\widehat{R}_N(\theta) := \frac{1}{N}\sum_{i=1}^N \ell(\theta;\mathbf{Z}_0^{(i)})$. The estimator $\widehat{\theta}$ minimizes $\widehat{R}_N(\theta)$ over the compact set $\Theta \subset \mathbb{R}^p$.

Because $p$ is finite and the loss is bounded or sub-Gaussian, classical parametric uniform convergence guarantees apply. Concretely, by, for example, the standard Glivenko–Cantelli or Vapnik–Chervonenkis theory for finite-dimensional parameter spaces (see, e.g., \citet{shalev2014understanding} or \citet{geer2000empirical}), there is a function $\epsilon(N,\delta)$ going to $0$ as $N\to\infty$ such that, with probability at least $1-\delta$, one has
\[\sup_{\theta \in \Theta} \Bigl|\, \widehat{R}_N(\theta) - R(\theta)\Bigr| \;\;\le\; \epsilon(N,\delta).\]
Since $\widehat{\theta}$ minimizes $\widehat{R}_N(\theta)$, a straightforward argument gives
\[ R(\widehat{\theta}) \;\;=\;\; \widehat{R}_N(\widehat{\theta}) \;+\;\bigl(R(\widehat{\theta})-\widehat{R}_N(\widehat{\theta})\bigr) \;\;\le\;\; \widehat{R}_N\!\bigl(\theta^*\bigr) \;+\;\epsilon(N,\delta),\]
where $\theta^* \in \arg\min_{\theta \in \Theta} R(\theta)$. Another $\epsilon(N,\delta)$-term bounds the difference between $\widehat{R}_N(\theta^*)$ and $R(\theta^*)$, which yields
\[R(\widehat{\theta}) \;\;\le\;\; R(\theta^*) \;+\; 2\,\epsilon(N,\delta).\]
Thus, possibly redefining $\epsilon(N,\delta)$ by a small constant factor, we obtain the usual generalization guarantee
\[R(\widehat{\theta}) \;\;\le\;\; \min_{\theta\in\Theta} R(\theta) \;+\;\epsilon(N,\delta).\]
Because $R(\theta)$ is exactly the expected in-context loss in question, this completes the proof.
\end{proof}

\textbf{\textit{Note.}} The boundedness (or sub-Gaussianity) of $\ell(\theta;\mathbf{Z}_0)$, plus the compactness and continuity assumptions on $\theta$, are key to ensuring uniform convergence in finite-dimensional parameter spaces. Such conditions are typically satisfied in standard LFOM settings, e.g.\ when the data are bounded or sub-Gaussian and the square loss is used. If data or parameters are unbounded, one can impose appropriate norm constraints and regularity conditions to obtain similar results.

\section{Comparison to Nesterov Accelerated Gradient Method (NAG) and Momentum Gradient Descent (MGD)}

\subsection{Nesterov Accelerated Gradient Method (NAG)}

NAG is a commonly used optimization technique that builds on classical gradient descent by incorporating a momentum term that anticipates the next update. Specifically, the weights are updated using the following update rules:
\[
\mathbf{v}_{k+1} = \mathbf{w}_k + \beta_k (\mathbf{w}_k - \mathbf{w}_{k-1})
\]
\[
\mathbf{w}_{k+1} = \mathbf{v}_{k+1} - \eta_k \nabla f(\mathbf{v}_{k+1})
\]

Here, \( \beta_k \) controls the influence of previous updates (momentum), and \( \eta_k \) is the learning rate. In our experiments, we selected \( \eta_k = 0.03 \) and \( \beta_k = 0.9 \) after testing various values of these parameters on the given distribution, as in Section \ref{Section:Experimental_Results}. These values provided the best performance. The momentum term allows NAG to ``look ahead" in the optimization trajectory, which often leads to faster convergence than vanilla gradient descent.

\subsection{Momentum Gradient Descent (MGD)}

Momentum Gradient Descent operates similarly to NAG but without the anticipation of future steps. The algorithm updates the weights based on a momentum term that accelerates convergence in directions with consistent gradients. The update rule for MGD is given by:
\[
\mathbf{v}_{k+1} = \beta_k \mathbf{v}_k - \eta_k \nabla f(\mathbf{w}_k)
\]
\[
\mathbf{w}_{k+1} = \mathbf{w}_k + \mathbf{v}_{k+1}
\]

In our experiments, the learning rate \( \eta_k = 0.005 \) and momentum parameter \( \beta_k = 0.9 \) provided the best results on the given distribution, as in Section \ref{Section:Experimental_Results}. Momentum helps to mitigate oscillations in directions with high curvature, stabilizing the optimization trajectory and leading to faster convergence compared to gradient descent.

\subsection{Memformers vs. NAG and MGD}

\begin{figure}[htbp]
  \centering
  \begin{minipage}[b]{0.45\textwidth}
    \centering
    \includegraphics[width=\textwidth]{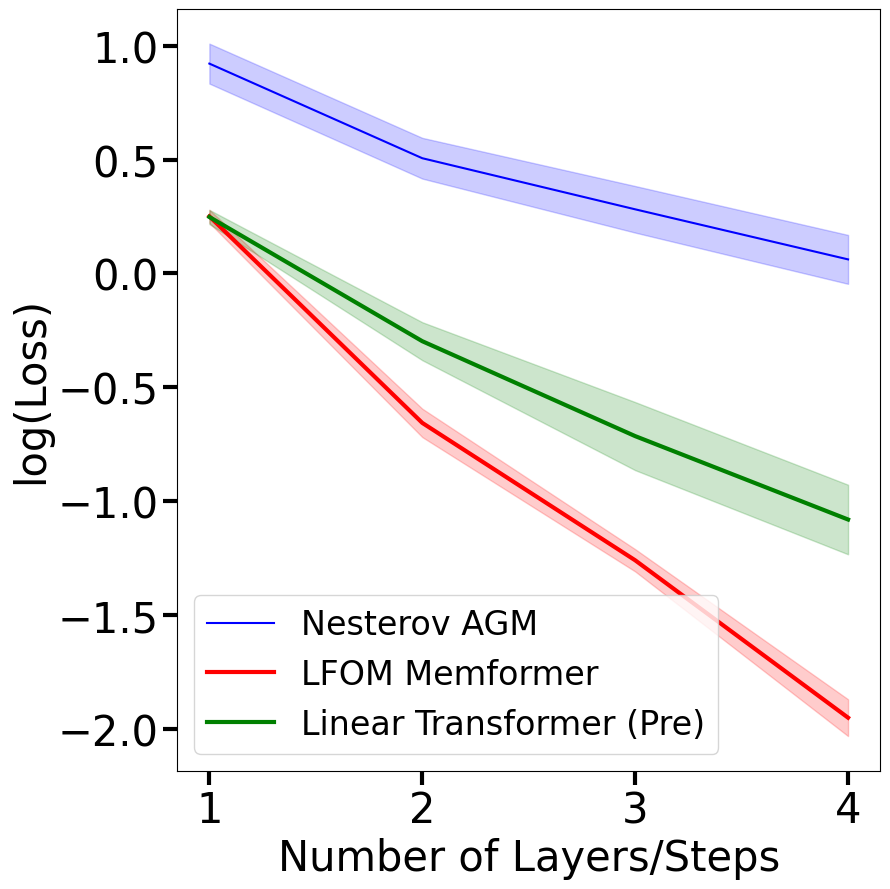}
    \caption{Nesterov AGM vs. LFOM Memformer}
    \label{fig:nag_vs_Memformer}
  \end{minipage}
  \hfill
  \begin{minipage}[b]{0.45\textwidth}
    \centering
    \includegraphics[width=\textwidth]{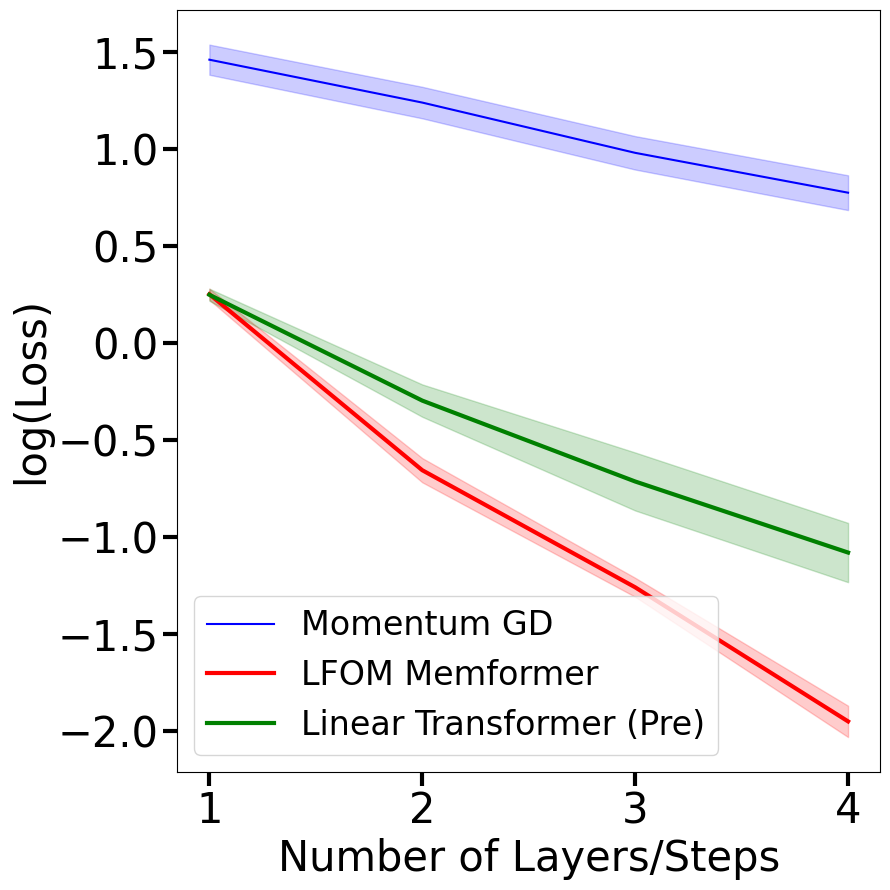}
    \caption{Momentum GD vs. LFOM Memformer}
    \label{fig:mgd_vs_Memformer}
  \end{minipage}
  \caption{Comparison of LFOM Memformer with (a) Nesterov AGM and (b) Momentum GD.}
  \label{fig:comparison_figures}
\end{figure}

In our experiments, we observed that Memformers \eqref{LFOM_memory} outperform both NAG and MGD. Figures~\ref{fig:nag_vs_Memformer} and \ref{fig:mgd_vs_Memformer} compare the performance of Memformer with NAG and MGD, respectively, on the same non-isotropic data. As shown, the Memformer achieves faster convergence and much better loss performance compared to both algorithms.

\section{LFOM Memformer Experiments With More Than 4 Layers}

In our experiments, we observed that Memformers with more than 4 layers continue to demonstrate impressive performance in learning optimization strategies. We conducted experiments with Memformers having up to 7 layers and dimension \( d = 10 \). Training beyond this point becomes impractical due to extensive iteration requirements and significant convergence times, which can span several hours. This limitation is a consequence of computational constraints (e.g., available GPUs) rather than any inherent deficiency of the Memformer architecture itself. Here, \( d \) refers to the rank of the square matrix \( \mathbf{X} \mathbf{X}^T \) in the empirical loss quadratic as described in Equation \ref{in-context-loss}.

\begin{figure}[htbp]
  \centering
  \begin{minipage}[b]{0.48\textwidth}
    \centering
    \includegraphics[width=\textwidth]{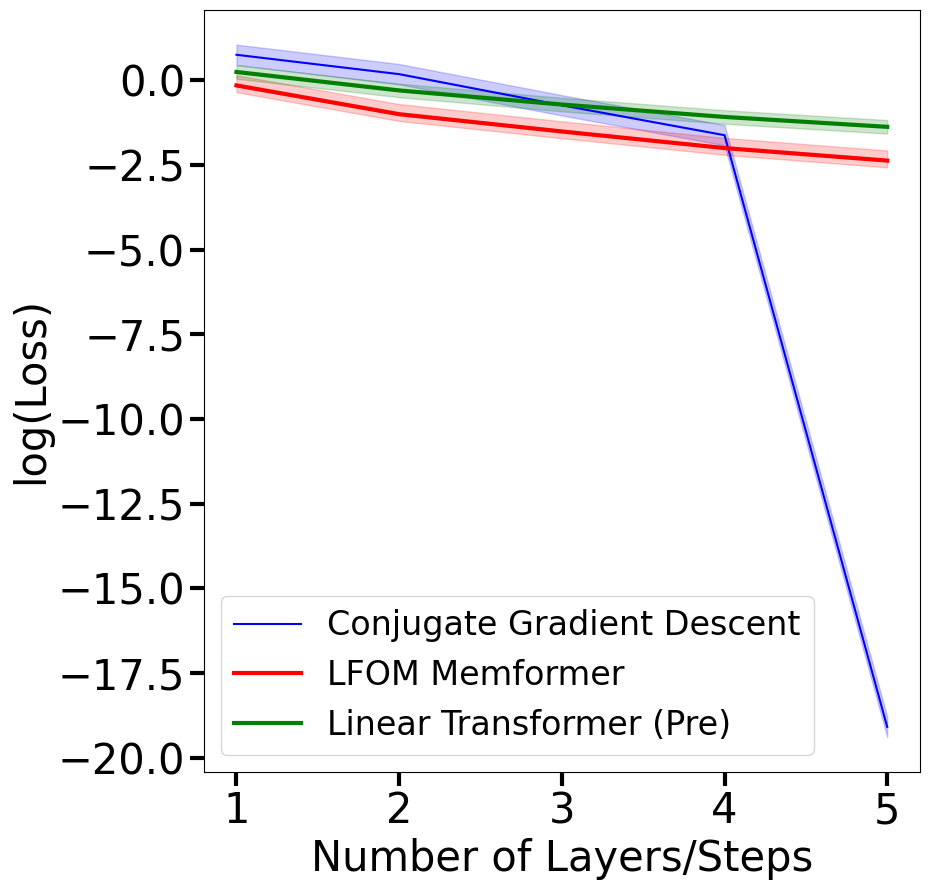}
    \caption{Memformer performance for \( d = 5 \) with 5 layers.}
    \label{fig:layers5}
  \end{minipage}
  \hfill
  \begin{minipage}[b]{0.48\textwidth}
    \centering
    \includegraphics[width=\textwidth]{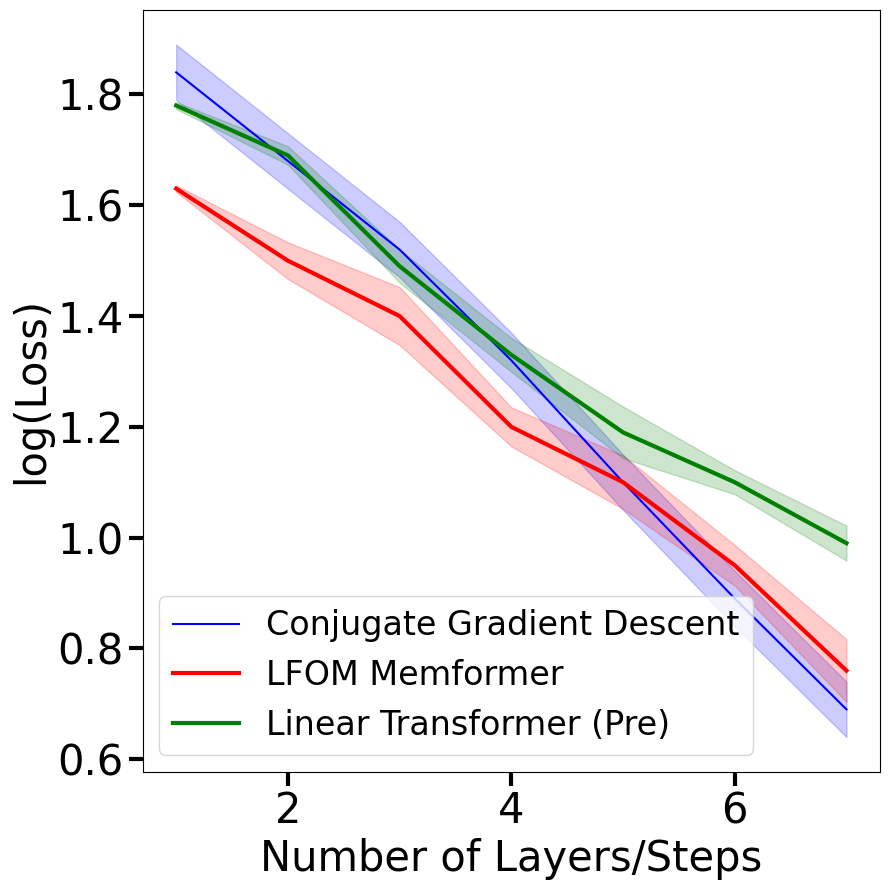}
    \caption{Memformer performance for \( d = 10 \) with 7 layers.}
    \label{fig:layers7}
  \end{minipage}
  \caption{Comparison of Memformer performance for different configurations of depth \( d \) and layers.}
  \label{fig:layers_comparison}
\end{figure}

1. \textbf{Experiment (Figure \ref{fig:layers5})} (Dimension \( d = 5 \), Layers = 5): As expected, Conjugate Gradient Descent (CGD) converges within \( d \) steps due to the dimensionality constraint. Remarkably, even though the Memformer only learns general parameters \( \mathbf{A}_\ell \) (Equation~\eqref{params_Thm3}) and \( \Gamma_\ell \) (Equation~\eqref{LFOM_memory}), it manages to keep up with CGD for up to 4 steps, showcasing its efficiency.

2. \textbf{Experiment (Figure \ref{fig:layers7})} (Dimension \( d = 10 \), Layers = 7): In this case, CGD does not converge until beyond 7 steps, which aligns with theoretical expectations. Nevertheless, the Memformer remains highly competitive, matching CGD's performance for 6 steps and even performing comparably at 7 steps. This demonstrates the Memformer’s robust generalization capabilities, even under more complex conditions.

\section{Experiment on Convergence Verification for Memformer Parameter \( \mathbf{A}_\ell \) to \( \Sigma \)}

Our strategy to train the Memformer \eqref{LFOM_memory} was to first train the \( A_\ell \)'s \eqref{params_Thm3} in each layer \(\ell\) on the training batch and then to ``fine-tune" the \( \Gamma_\ell \)'s on the training batch. Therefore, we present here an empirical verification of our results per \textbf{Theorem 3} in \cite{ahn2024transformers}. 

\textbf{Theorem 3. (\cite{ahn2024transformers})} \textit{Assume that \( x^{(i)} \overset{\text{iid}}{\sim} \mathcal{N}(0, \Sigma) \) and \( w_x \sim \mathcal{N}(0, \Sigma^{-1}) \), for \( i = 1, \ldots, n \), and for some \( \Sigma \succ 0 \). Consider the optimization of in-context loss \eqref{in-context_f} for a \( k \)-layer transformer with the parameter configuration in Eq. \eqref{params_Thm3} given by:
\[
\min_{\{A_\ell\}_{\ell=0}^{L-1}} f(A).
\]
Let \( S \subset \mathbb{R}^{L \times d \times d} \) be defined as follows: \( A \in S \) if and only if for all \( i = 0, \ldots, L-1 \), there exist scalars \( a_i \in \mathbb{R} \) such that \( A_i = a_i \Sigma^{-1} \). Then
\[
\inf_{(A, B) \in S} \sum_{i=0}^{L-1} \|\nabla_{A_i} f(A, B)\|_F^2 = 0,
\]
where \( \nabla_{A_i} f \) denotes the derivative with respect to the Frobenius norm \( \|A_i\|_F \).}

We evaluated the in-context learning (ICL) loss for linear regression with \( d = 5 \) and \( n = 20 \), where \( x^{(i)} \sim \mathcal{N}(0, \Sigma) \) and \( w_x \sim \mathcal{N}(0, \Sigma^{-1}) \). The covariance \( \Sigma \) was generated as \( \Sigma = U^T D U \), with \( U \) being a random orthogonal matrix and \( D = \text{diag}(1, 1, 1/4, 1/16, 1) \). A three-layer linear transformer was trained using ADAM, with \( A_0, A_1, A_2 \) initialized as i.i.d. Gaussian matrices. Each gradient step used minibatches of size 20,000, resampled every 100 steps, and gradients were clipped to 0.01. Results were averaged over 5 runs with independent \( U \) and \( \Sigma \) samples.

\begin{figure}[htbp]
  \centering
  \begin{minipage}[b]{0.32\textwidth}
    \centering
    \includegraphics[width=\textwidth]{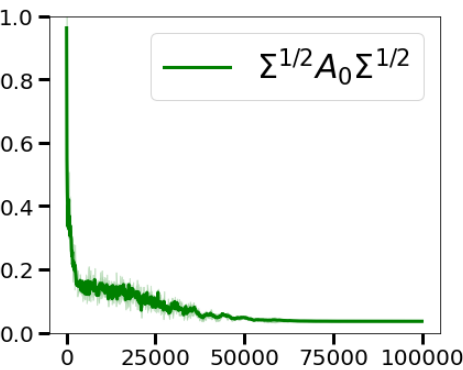}
    \caption{\( A_0 \) convergence.}
    \label{fig:sigma_a0}
  \end{minipage}
  \hfill
  \begin{minipage}[b]{0.32\textwidth}
    \centering
    \includegraphics[width=\textwidth]{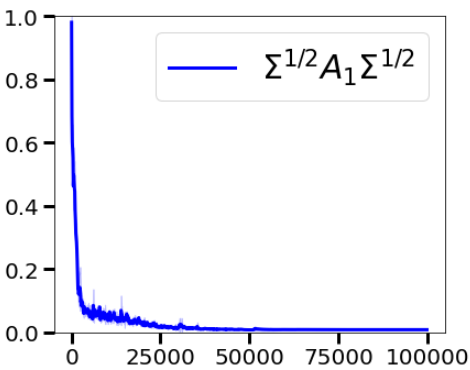}
    \caption{\( A_1 \) convergence.}
    \label{fig:sigma_a1}
  \end{minipage}
  \hfill
  \begin{minipage}[b]{0.32\textwidth}
    \centering
    \includegraphics[width=\textwidth]{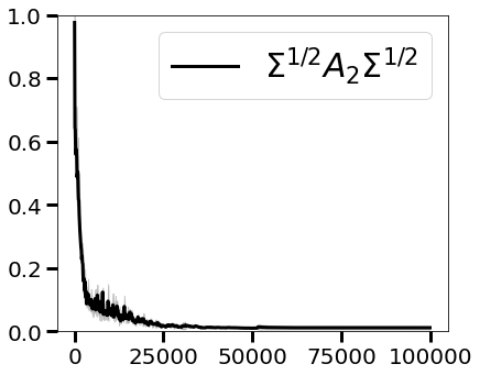}
    \caption{\( A_2 \) convergence.}
    \label{fig:sigma_a2}
  \end{minipage}
  \caption{Convergence comparison of \( A_0 \), \( A_1 \), and \( A_2 \).}
  \label{fig:sigma_comparison}
\end{figure}

To measure convergence, we computed the normalized Frobenius norm distance:
\[
\text{Dist}(M, I) := \min_{\alpha} \frac{\| M - \alpha I \|_F}{\| M \|_F},
\]
where
\[
\alpha := \frac{1}{d} \sum_{i=1}^d M[i, i],
\]
which quantifies the deviation of \( M / \| M \|_F \) from a scaled identity. The distance \(\text{Dist}(\Sigma^{1/2} A_i \Sigma^{1/2}, I)\), averaged over 5 runs, is shown in Figures \ref{fig:sigma_a0}, \ref{fig:sigma_a1}, and \ref{fig:sigma_a2} as a function of training iterations.

\end{document}